\documentclass[10pt,twocolumn,letterpaper]{article}

\usepackage[pagenumbers]{cvpr} 

\usepackage[accsupp]{axessibility}
\usepackage{graphicx}
\usepackage{textcomp}
\usepackage{amsmath}
\usepackage{amssymb}
\usepackage{booktabs}
\usepackage{multirow}
\usepackage{enumitem}
\usepackage{textcomp}
\usepackage{algorithm}
\usepackage{algorithmicx}
\usepackage{algpseudocode}

\usepackage{bm}
\usepackage{bbm}
\usepackage{indentfirst}
\usepackage{url}
\usepackage[table,dvipsnames]{xcolor}
\setlist[itemize]{noitemsep, nolistsep}
\usepackage[pagebackref,breaklinks,colorlinks]{hyperref}

\usepackage[capitalize]{cleveref}
\crefname{section}{Sec.}{Secs.}
\Crefname{section}{Section}{Sections}
\Crefname{table}{Table}{Tables}
\crefname{table}{Tab.}{Tabs.}

\newcommand{\mystd}[1]{\footnotesize{ \textpm{ #1}}}

\DeclareMathOperator*{\argmax}{arg\,max}

\def\cvprPaperID{7376} 
\def\confName{ICCV}
\def\confYear{2023}

\begin{document}

\title{IOMatch: Simplifying Open-Set Semi-Supervised Learning \\ with Joint Inliers and Outliers Utilization}

\author{
Zekun~Li$^{1}$ \quad Lei Qi$^2$ \quad Yinghuan Shi$^{1,}$\thanks{Corresponding author: Yinghuan Shi (syh@nju.edu.cn). Zekun Li, Yinghuan Shi and Yang Gao are with the State Key Laboratory for Novel Software Technology and National Institute of Healthcare Data Science, Nanjing University. Lei Qi is with the School of Computer Science and Engineering, Southeast University. This work is supported by NSFC Program (62222604, 62206052, 62192783), China Postdoctoral Science Foundation Project (2023T160100), Jiangsu Natural Science Foundation Project (BK20210224), and CCF-Lenovo Bule Ocean Research Fund.} \quad Yang Gao$^{1}$ \\
$^1$~Nanjing University \quad $^2$ Southeast University \\
}

\maketitle

\begin{abstract}
   Semi-supervised learning (SSL) aims to leverage massive unlabeled data when labels are expensive to obtain. Unfortunately, in many real-world applications, the collected unlabeled data will inevitably contain unseen-class outliers not belonging to any of the labeled classes. To deal with the challenging open-set SSL task, the mainstream methods tend to first detect outliers and then filter them out. However, we observe a surprising fact that such approach could result in more severe performance degradation when labels are extremely scarce, as the unreliable outlier detector may wrongly exclude a considerable portion of valuable inliers. To tackle with this issue, we introduce a novel open-set SSL framework, IOMatch, which can jointly utilize inliers and outliers, even when it is difficult to distinguish exactly between them. Specifically, we propose to employ a multi-binary classifier in combination with the standard closed-set classifier for producing unified open-set classification targets, which regard all outliers as a single new class. By adopting these targets as open-set pseudo-labels, we optimize an open-set classifier with all unlabeled samples including both inliers and outliers. Extensive experiments have shown that IOMatch significantly outperforms the baseline methods across different benchmark datasets and different settings despite its remarkable simplicity. Our code and models are available at \href{https://github.com/nukezil/IOMatch}{\small\url{https://github.com/nukezil/IOMatch}}.
\end{abstract}

\vspace{-5mm}
 
\section{Introduction}
\label{sec:intro}
Semi-supervised learning (SSL) \cite{ChaepelleBook} is a classical machine learning paradigm that attempts to improve a model's performance by utilizing unlabeled data in addition to insufficient labeled data. With a tiny fraction of labeled data, advanced deep SSL methods can achieve the performance of fully supervised methods in some cases, such as image classification \cite{sohn2020fixmatch} and semantic segmentation \cite{yang2021st++}. 

Most existing SSL methods rely on the fundamental assumption that labeled and unlabeled data share the same class space. However, it is usually difficult, even impossible, to collect such a unlabeled data set in many real-world applications since we can not manually examine the massive unlabeled data. Therefore, a more challenging scenario arises, where unseen-class outliers not belonging to any of the labeled classes exist in the unlabeled data. Such setting is called Open-Set Semi-Supervised Learning (OSSL) \cite{yu2020multi}.

\begin{figure}[t]
  \centering
  \includegraphics[width=\linewidth]{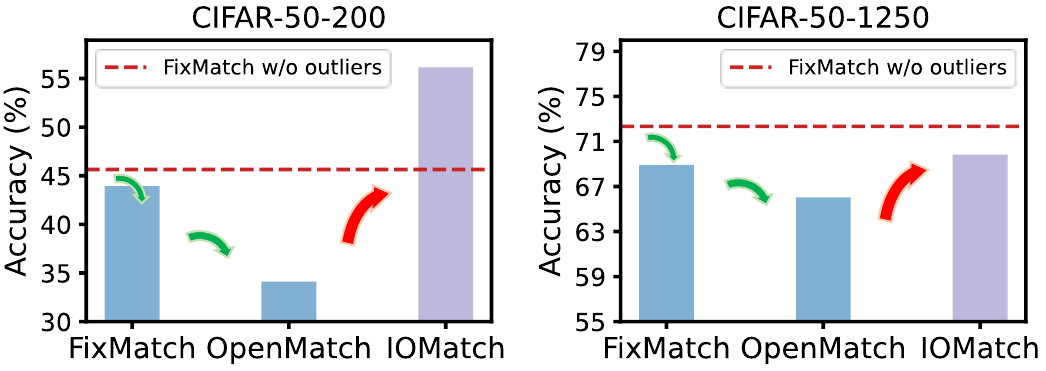}
  \vspace{-0.65cm}
  \caption{The motivation of our work comes from a surprising fact in open-set semi-supervised learning tasks: An unreliable outlier detector can be more harmful than outliers themselves, because it will wrongly exclude valuable inliers from subsequent training. For this issue, we consider a unified paradigm for utilizing open-set unlabeled data, even when it is difficult to distinguish exactly between inliers and outliers, and thus we propose IOMatch.}
  \label{fig:motivation}
  \vspace{-0.6cm}
\end{figure}

The negative effects of unseen-class outliers have been observed in a pioneer work \cite{oliver2018realistic}. As the research of SSL has grown rapidly in recent years, we extensively evaluate more advanced SSL methods. Some of the key results are shown in Figure \ref{fig:motivation}, in which we plot the performance under standard and open-set SSL as the dash lines and charts, respectively. Taking the classical method, FixMatch \cite{sohn2020fixmatch}, as an example, we can observe that adding extra outliers does hurt the classification accuracy compared to the standard SSL setting with no outlier, because it is impossible to obtain correct seen-class pseudo-labels for these outliers. An intuitive approach to handle outliers is to detect and remove them, as OpenMatch \cite{saito2021openmatch} does. In particular, it combines FixMatch with an outlier detector. The detector is first pre-trained and then used to retain only inliers for FixMatch training. However, we find that such approach actually results in more severe performance degradation especially when labels are extremely scarce. The reason is that the pre-trained detector can be so unreliable that it will wrongly exclude a considerable portion of valuable inliers from subsequent training. In this regard, a surprising fact is that \textit{a bad detector is worse than no detector at all}. Similar to OpenMatch, other existing methods \cite{yu2020multi, huang2021trash, he2022safe} using various outlier detectors also suffer from this issue, as they all follow the detect-and-filter paradigm.

\begin{figure}[t]
  \centering
  \includegraphics[width=0.97\linewidth]{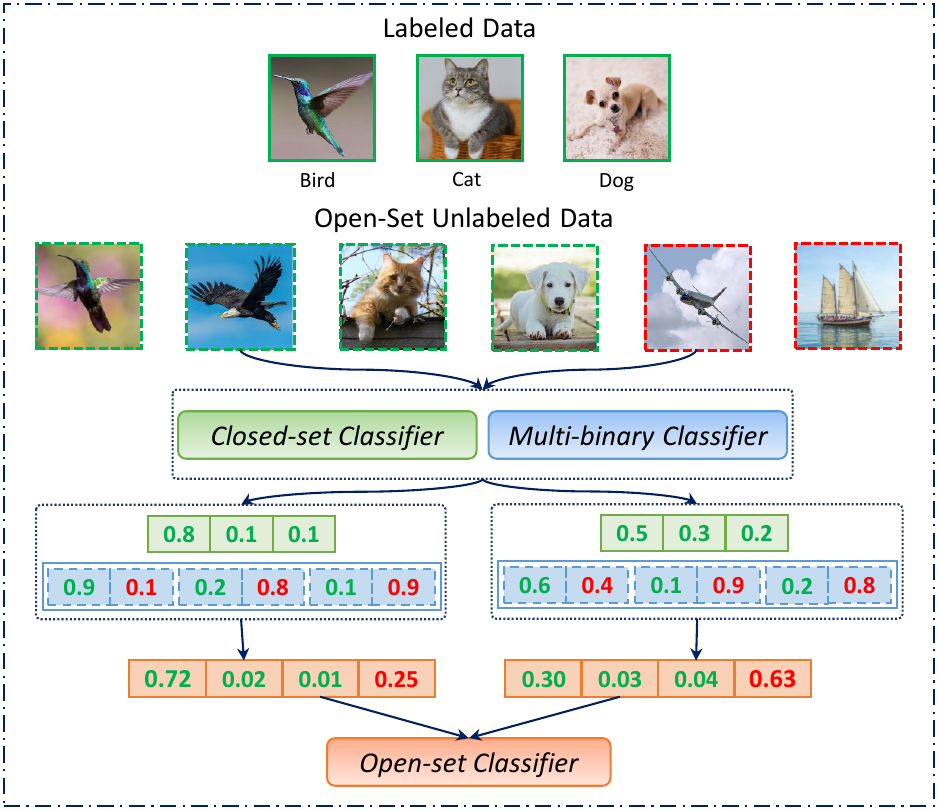}
  \vspace{-0.3cm}
  \caption{Illustration of joint inliers and outliers utilization. We fuse the predictions of the closed-set classifier and the multi-binary classifier to produce the open-set targets for both inliers and outliers, where outliers are regarded as a single new class (denoted in red). All the open-set unlabeled data will be fully exploited by optimizing an open-set classifier via pseudo-labeling.}
  \label{fig:idea}
  \vspace{-0.4cm}
\end{figure}

From the above analysis, we can observe that the performance of existing OSSL methods is highly dependent on the unseen-class detection. However, it is difficult indeed to obtain a reliable outlier detector in the early stage of training due to the scarcity of labels. Thus, instead of sending open-set unlabeled samples into different learning branches (\eg, inliers for pseudo-labeling and outliers being thrown away), we are better to deal with them in a unified paradigm. This allows the opportunity to make corrections even if the unseen-class detection is not accurate at the beginning.

In this paper, we consider a novel strategy to jointly utilize inliers and outliers without distinguishing exactly between them, and thus propose a simple yet effective OSSL framework, IOMatch. Along with the standard closed-set classifier, IOMatch adopts a multi-binary classifier \cite{saito2021ovanet} that predicts how likely a sample is to be an inlier of each seen class. We fuse the predictions of the two classifiers to produce unified open-set classification targets by regarding all outliers as a new class. These open-set targets are then utilized to train an open-set classifier with both unlabeled inliers and outliers via pseudo-labeling. We illustrate the core idea in Figure \ref{fig:idea}. Different from the detect-and-filter methods \cite{yu2020multi, huang2021trash, saito2021openmatch, he2022safe}, all the network modules of IOMatch are simultaneously optimized, which makes it easy to use.

\begin{figure}[t]
  \centering
  \includegraphics[width=\linewidth]{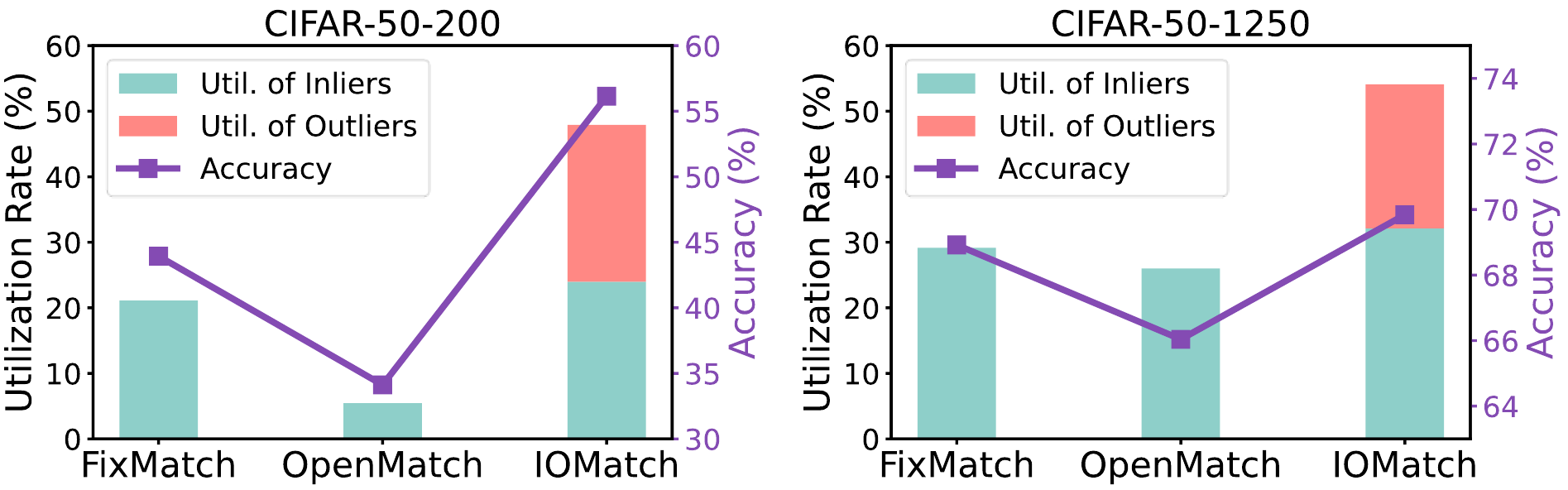}
  \vspace{-0.7cm}
  \caption{We define the utilization rate of open-set unlabeled data as the ratio of selected correct pseudo-labels to all unlabeled samples. Compared to the previous methods, IOMatch can not only retain more valuable inliers but also utilize additional outliers by adopting open-set targets as pseudo-labels.}
  \label{fig:util_rate}
  \vspace{-0.4cm}
\end{figure}

We conduct extensive experiments to demonstrate the effectiveness of IOMatch across different benchmark datasets and different settings. The performance gains are significant especially when labels are scarce and class mismatch is severe. For instance, on the CIFAR-100 dataset, IOMatch outperforms the current state-of-the-art by 7.46\% and 4.78\%, when the proportion of outliers is as high as 80\% and 50\%, and only 4 labels  per seen class are available. Figure \ref{fig:util_rate} explains why IOMatch is able to achieve such improvements: Compared to the existing OSSL methods, IOMatch avoids incorrect exclusion of valuable inliers; Compared to the standard SSL methods, IOMatch can additionally utilize ``poisonous'' outliers. In a nutshell, with the novel paradigm of joint inliers and outliers utilization, open-set unlabeled data can be more fully exploited by IOMatch.

We summarize our contributions as follows:
\begin{itemize}[itemsep=2pt,topsep=2pt]
\item We reveal that existing open-set SSL methods could easily fail due to their unreliable outlier detectors when labels are extremely scarce.
\item We propose a novel open-set SSL framework called IOMatch that can jointly utilize both inliers and outliers in a unified paradigm.
\item We perform comprehensive experiments across various OSSL settings. In spite of its simplicity, IOMatch significantly outperforms the strong rivals, especially when the tasks are challenging.
\end{itemize}

\section{Related Work}
\label{sec:relat}
\subsection{Semi-Supervised Learning}
For mainstream deep SSL approaches, consistency regularization \cite{bachman2014learning} is a crucial technique and has been widely adopted in many works \cite{sajjadi2016regularization,laine2016temporal,tarvainen2017mean,miyato2018virtual,berthelot2019mixmatch}. Briefly speaking, this technique enforces the model to output a consistent prediction on the different perturbed versions of the same sample. Among existing works, FixMatch \cite{sohn2020fixmatch} is one of the most influential SSL methods, which is popular for its simplicity and effectiveness. It improves consistency regularization with strong data augmentation and performs pseudo-labeling based on confidence thresholding. There are many other works that have made important technical contributions to the research of SSL. ReMixMatch \cite{berthelot2019remixmatch} introduces distribution alignment and augmentation anchoring. FlexMatch \cite{flexmatch} and FreeMatch \cite{wang2022freematch} propose to adjust the class-specific confidence thresholds based on the different learning difficulties. CoMatch \cite{li2021comatch} and SimMatch \cite{zheng2022simmatch} incorporate contrastive learning objectives to exploit instance-level similarity. More comprehensive reviews on SSL theories and methods can be found in \cite{van2020survey,yang2022survey,usb2022}.

Despite the remarkable success on various SSL tasks, all these methods assume that labeled and unlabeled data share the same class space. Such assumption could be difficult to satisfy in real-world applications, which may lead to considerable performance degradation. Therefore, it is necessary to consider the more practical open-set SSL setting. 

\subsection{Open-Set Semi-Supervised Learning}
As the standard closed-set classifier cannot assign correct seen-class pseudo-labels for unseen-class outliers, an intuitive approach is to detect outliers and filter them out before pseudo-labeling. Mainstream OSSL methods adopt such detect-and-exclude strategy to reduce the perturbation from outliers. For example, UASD \cite{chen2020semi} considers the predictions of the closed-set classifier and use the confidence to identify outliers. Also with the predictions, SAFE-STUDENT \cite{he2022safe} defines an energy-discrepancy score to replace the confidence. There are other several methods which introduce additional network modules for unseen-class detection. MTCF \cite{yu2020multi} adopts a binary classification head which is trained in noisy label optimization paradigm. T2T \cite{huang2021trash} proposes a cross-modal matching module to predict whether a sample is matched to an assigned one-hot seen-class label. With the similar idea, OpenMatch \cite{saito2021openmatch} employs a group of one-vs-all classifiers as the outlier detector. 

Although the above OSSL methods are effective when labels are relatively sufficient (\eg, 100 labels per seen class or more), it is hard to achieve satisfactory unseen-class detection performance when the number of labeled samples is extremely limited. In such a challenging scenario, even after a pre-training stage, the outlier detector still does not perform well due to the scarcity of labels. As a consequence, it will tend to wrongly exclude a large portion of unlabeled inliers. Without exposure to these misidentified samples, such errors are quite difficult to rectify, which will lead to more severe performance degradation than that caused by outliers themselves. A few recent methods propose to perform extra pretext tasks, such as rotation recognition \cite{huang2021trash} and label distribution calibration \cite{he2022safe}, with the detected outliers. These techniques may mitigate the adverse affects of the unreliable outlier detector, but cannot really address the issue. 

Another related learning problem is out-of-distribution (OOD) detection \cite{hendrycks2016baseline}, which aims to separate OOD samples from in-distribution (ID) samples. OOD detection has different problem formulation and learning objectives from OSSL, so it is out of the scope of this work. For further discussions about the connections and differences between the two problems, please refer to the supplementary material.

\section{IOMatch}
\label{sec:method}
\subsection{Preliminaries and Overview}
We define the open-set semi-supervised learning task as following. For a $K$-class classification problem, let $\mathcal{X}=\{(\bm{x}_i,y_i):i\in(1,\dots,B)\}$ be a batch of $B$ labeled samples, where $\bm{x}_i$ is a training sample and $y_i\in \{1,\dots,K\}$ is the corresponding label. Let $\mathcal{U}=\{\bm{u}_i:i\in (1,\dots,\mu B)\}$ be a batch of $\mu B$ unlabeled samples, where $\mu$ is a hyperparameter that determines the relative sizes of $\mathcal{X}$ and $\mathcal{U}$. In the OSSL task, there exists a subset  $\mathcal{U}^{out}\subset \mathcal{U}$, where $\mathcal{U}^{out}=\{\bm{u}^{out}\}$ and $\bm{u}^{out}$ does not belong to any of the $K$ seen classes. Then, $\mathcal{U}^{out}$ are called unseen-class \textit{outliers} and the rest of unlabeled samples $\mathcal{U}^{in}=\mathcal{U}/\mathcal{U}^{out}$ are called seen-class \textit{inliers}. 

Given a labeled batch $\mathcal{X}$, we apply a random weak transformation function $\mathcal{T}_w(\cdot)$ to obtain the weakly augmented samples.  A base encoder network $f(\cdot)$ is employed to extract the features from these samples, \ie, $\bm{h}_i=f(\mathcal{T}_w(\bm{x}_i))\in \mathbb{R}^D$. A closed-set classifier $\phi(\cdot)$ maps the feature $\bm{h}_i$ into the predicted seen-class probability distribution, \ie, $\bm{p}_i=\phi(\bm{h}_i)$. The labeled batch are used to optimize the networks with the standard cross-entropy loss $\text{H}(\cdot,\cdot)$:
\begin{equation}
\label{eq:loss-s}
    \mathcal{L}_{s}(\mathcal{X})=\frac{1}{B}\sum_{i=1}^B \text{H}(y_i,\bm{p}_i).
\end{equation} 
Additionally, we adopt a projection head $g(\cdot)$ to obtain the low-dimensional embedding $\bm{z}_i=g(\bm{h}_i)\in \mathbb{R}^d$ and then a multi-binary classifier $\chi(\cdot)$ to produce the class-wise likelihood of inliers or outliers $\bm{o}_i=\chi(\bm{z}_i)\in\mathbb{R}^{2K}$.  

For an unlabeled batch $\mathcal{U}$, we apply both the weak and strong augmentation with $\mathcal{T}_w(\cdot)$ and $\mathcal{T}_s(\cdot)$. The same operations as above are performed to obtain $\bm{h}^w_i$, $\bm{z}^w_i$, $\bm{p}^w_i$, and $\bm{o}^w_i$ for the weakly augmented samples $\mathcal{T}_w(\bm{u}_i)$; $\bm{h}^s_i$, $\bm{z}^s_i$, $\bm{p}^s_i$ and $\bm{o}^s_i$ for the strongly augmented samples $\mathcal{T}_s(\bm{u}_i)$. Moreover, an open-set classifier $\psi(\cdot)$ is introduced for the unlabeled samples to predict the open-set probability distribution, where all outliers are regarded as a single new class.

\begin{figure}[t]
  \centering
  \includegraphics[width=0.95\linewidth]{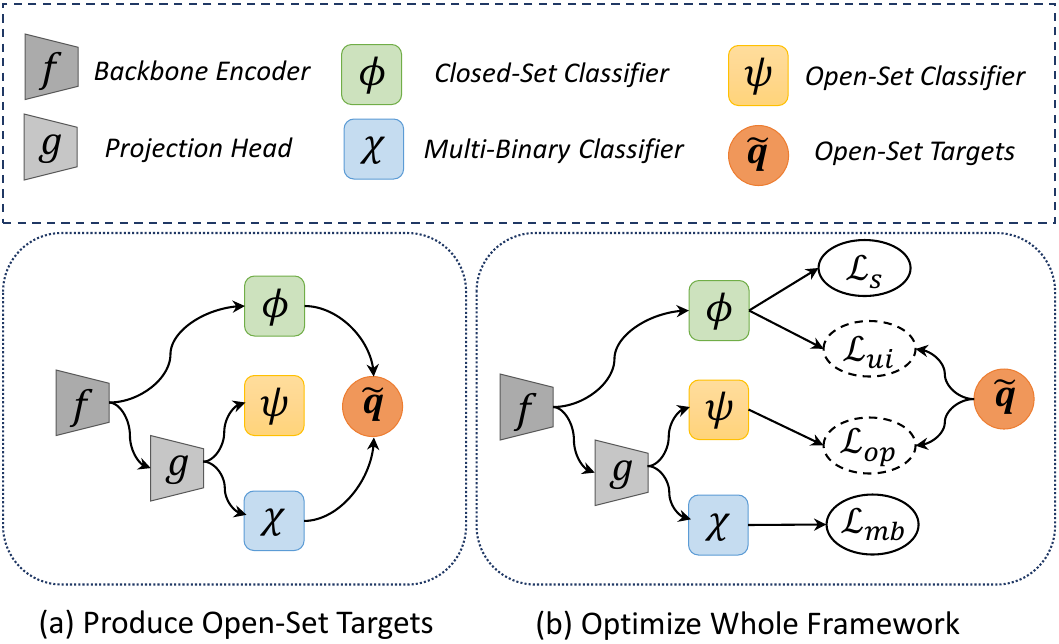}
  \vspace{-2mm}
  \caption{Overview of IOMatch. In each iteration, we first employ the closed-set classifier and the multi-binary classifier to produce the open-set targets, which are then used for selecting high-quality inliers and utilizing outliers. All the network modules in IOMatch are simultaneously optimized with four learning objectives.} 
  \label{fig:framework}
  \vspace{-5mm}
\end{figure}

The overall framework of IOMatch is illustrated in Figure \ref{fig:framework}. We propose a novel approach to produce unified open-set targets by fusing predictions of the closed-set classifier and the multi-binary classifier. These targets are then used to optimize the closed-set and open-set classifiers to achieve joint inliers and outliers utilization. As an one-stage method, IOMatch shows remarkable simplicity and is easy to deploy across various OSSL settings.

\subsection{Unified Open-Set Targets Production}
As the standard closed-set classifier can only assign each sample to one of the seen classes, we employ an additional multi-binary classifier which has been proved capable in related unseen-class detection problems \cite{saito2021ovanet,saito2021openmatch,zhu2022crossmatch}. The multi-binary classifier can be viewed as a combination of $K$ sub-classifiers, \ie, $\chi=\{\chi_k:k\in\ (1,\dots,K)\}$. Technically, $\chi_k$ is the binary classifier for the $k$-th seen class with the output $\bm{o}_{i,k}=\chi_k(\bm{z}_i)\in \mathbb{R}^2$, where $\bm{o}_{i,k}=(o_{i,k}, \bar{o}_{i,k})$ and $o_{i,k}+\bar{o}_{i,k}=1$. $\bm{o}_{i,k}$ is a probability distribution to indicate how likely the sample $\bm{x}_i$ is to be an inlier or an outlier with respect to the $k$-th seen class. The hard-negative sampling strategy \cite{saito2021ovanet} is adopted to optimize the multi-binary classifier with the labeled samples:
\begin{equation}
\label{eq:loss-mb}
    \mathcal{L}_{mb}(\mathcal{X})=\frac{1}{B}\sum_{i=1}^B \left( -\log(o_{i,y_i})-\min_{k\neq y_i}\log(\bar{o}_{i,k})\right).
\end{equation}

Combining the multi-binary classifier with the closed-set classifier makes it possible to identify outliers. In the previous work, OpenMatch \cite{saito2021openmatch}, an unlabeled sample $\bm{u}_i$ is first assigned to one of the $K$ seen classes according to the closed-set prediction, \ie, $\hat{y_i}=\argmax_k (p^w_{i,k})$. Then, the binary probability $o^w_{i,\hat{y_i}}$ is considered to decide whether the sample is an inlier of the $\hat{y_i}$-th seen class or an unseen-class outlier, with the natural threshold of $0.5$. When the labels are relatively sufficient (\eg, 100 labels per class or more), such approach is effective, since the closed-set and the multi-binary classifiers can perform well after a pre-training stage with the labeled samples. However, when the number of labeled samples is limited, the one-hot pseudo-labels for seen classes will be hardly reliable.

Aware of this issue, we propose a novel approach to fully fuse the predictions of the two classifiers. Specifically, for each unlabeled sample $\bm{u}_i$, the seen-class probability distribution is predicted by the closed-set classifier, \ie, $\widetilde{\bm{p}}_i=\text{DA}(\phi(\bm{h}^w_i))$, where $\text{DA}(\cdot)$ stands for the distribution alignment strategy proposed by \cite{berthelot2019remixmatch} to balance the distribution of the model's predictions and thus prevent them from collapsing to certain classes. As the two classifiers are parameter-independent, $\widetilde{p}_{i,k}$ and $o^w_{i,k}$ are two distinct and complementary predictions on how likely $\bm{u}_i$ belongs to the $k$-th seen class. Therefore, for $1\le k \le K$, we use
\begin{equation}
    \widetilde{q}_{i,k} = \widetilde{p}_{i,k} \cdot o^w_{i,k}
\end{equation}
to estimate the probability that $\bm{u}_i$ belongs to the $k$-th seen class, when taking the possibility of outliers into consideration. Therefore, the probability that $\bm{u}_i$ is an outlier not belonging to any of the $K$ seen classes is estimated by
\begin{equation}
    \mathcal{S}_i = 1 - \sum_{j=1}^K  \widetilde{q}_{i,j} = \sum_{j=1}^{K} \widetilde{p}_{i,j} \cdot \bar{o}^w_{i,j}.
\end{equation}

Putting them all together produces a ($K$+1)-way class probability distribution $\widetilde{\bm{q}}_i\in \mathbb{R}^{K+1}$ by regarding all unseen classes as the virtual ($K$+1)-th class:
\begin{equation}
\label{eq:target-q}
    \widetilde{q}_{i,k} = 
    \begin{cases}
    \widetilde{p}_{i,k} \cdot o^w_{i,k}  & \text{if\quad} 1\leq k \leq K;
    \\[10pt] 
    \sum_{j=1}^{K} \widetilde{p}_{i,j} \cdot \bar{o}^w_{i,j} & \text{if\quad} k = K + 1. 
    \end{cases}
\end{equation}
In this way, we obtain a kind of unified open-set targets for all unlabeled samples, eliminating the need to precisely differentiate between inliers and outliers. This lays the foundation for the joint utilization of both inliers and outliers.

\begin{algorithm*}[t]
\caption{Optimization of IOMatch in Every Training Iteration}
\label{alg:algorithm}
\textbf{Input}: $\{(\bm{x}_i,y_i)\}_{i=1}^{B}$ and $\{\bm{u}_i\}_{i=1}^{\mu B}$: Labeled and unlabeled samples. $\mathcal{T}_w(\cdot)$ and $\mathcal{T}_s(\cdot)$: Weak and strong augmentation. $f(\cdot)$: Base encoder. $g(\cdot)$: Projection head. $\phi(\cdot)$: Closed-set classifier.  $\chi(\cdot)$: Multi-binary classifier. $\psi(\cdot)$: Open-set classifier. $\tau_p$ and $\tau_q$: Confidence thresholds. $\lambda_{mb}$, $\lambda_{ui}$, $\lambda_{op}$: Weights of losses. 
\begin{algorithmic}[1] 
\State $\bm{h}_i=f(\mathcal{T}_w(\bm{x}_i))$, $\bm{h}^w_i=f(\mathcal{T}_w(\bm{u}_i))$, $\bm{h}^s_i=f(\mathcal{T}_s(\bm{x}_i))$  \Comment{\textcolor{gray}{\textit{Obtain the features of the labeled and unlabeled samples.}}}
\State  $\bm{z}_i=g(\bm{h}_i)$, $\bm{z}^w_i=g(\bm{h}^w_i)$, $\bm{z}^s_i=g(\bm{h}^s_i)$ \Comment{\textcolor{gray}{\textit{Map the features into the projection space.}}} 
\State $\bm{p}=\phi(\bm{h}_i)$, $\widetilde{\bm{p}}=\text{DA}(\phi(\bm{h}^w_i))$, $\bm{p}^s=\phi(\bm{h}^s_i)$, $\bm{o}=\chi(\bm{z}_i)$, $\bm{o}^w=\chi(\bm{z}^w_i)$  \Comment{\textcolor{gray}{\textit{Make closed-set and multi-binary predictions.}}}
\State $\mathcal{L}_{s}(\mathcal{X})=\frac{1}{B}\sum_{i=1}^B \text{H}(y_i,\bm{p}_i)$ \Comment{\textcolor{gray}{\textit{Calculate the supervised loss.}}}
\State $\mathcal{L}_{mb}(\mathcal{X})=\frac{1}{B}\sum_{i=1}^B \left( -\log(o_{i,y_i})-\min_{k\neq y_i}\log(\bar{o}_{i,k})\right)$ \Comment{\textcolor{gray}{\textit{Calculate the multi-binary loss.}}}
\State $\widetilde{q}_{i,k} = \widetilde{p}_{i,k} \cdot o^w_{i,k} (1\leq k \leq K);$ $\widetilde{q}_{i,K+1} = \mathcal{S}_i = \sum_{j=1}^{K} \widetilde{p}_{i,j} \cdot \bar{o}^w_{i,j}$  \Comment{\textcolor{gray}{\textit{Produce open-set targets.}}}
\State $\mathcal{L}_{op}(\mathcal{U})=\frac{1}{\mu B}\sum_{i=1}^{\mu B} \mathbbm{1}(\max_k(\widetilde{q}_{i,k})>\tau_q) \cdot \text{H}(\bm{\widetilde{q}}_i,\bm{q}^s_i)$ \Comment{\textcolor{gray}{\textit{Calculate the open-set loss.}}}
\State $\mathcal{L}_{ui}(\mathcal{U})=\frac{1}{\mu B}\sum_{i=1}^{\mu B} \mathbbm{1}(\max_k(\widetilde{p}_{i,k})>\tau_p)\cdot \mathbbm{1}(\mathcal{S}_i < 0.5) \cdot \text{H}(\widetilde{\bm{p}}_i,\bm{p}^s_i)$ \Comment{\textcolor{gray}{\textit{Calculate the unlabeled inliers loss.}}}
\end{algorithmic}
\textbf{Output}: The overall loss $\mathcal{L}_{overall} = \mathcal{L}_s + \lambda_{mb}\mathcal{L}_{mb} + \lambda_{ui}\mathcal{L}_{ui} + \lambda_{op}\mathcal{L}_{op}$ to update the network parameters.
\end{algorithm*}

\subsection{Joint Inliers and Outliers Utilization}
For all the open-set unlabeled samples, we adopt the open-set targets as supervision to train the open-set classifier $\psi(\cdot)$ with its predictions $\bm{q}_i^s=\psi(\bm{z}^i_s)\in \mathbb{R}^{K+1}$ on the strongly augmented samples:
\begin{equation}
\label{eq:loss-op}
    \mathcal{L}_{op}(\mathcal{U})=\frac{1}{\mu B}\sum_{i=1}^{\mu B} \mathbbm{1}(\max_k(\widetilde{q}_{i,k})>\tau_q) \cdot \text{H}(\bm{\widetilde{q}}_i,\bm{q}^s_i),
\end{equation} 
where $\mathbbm{1}(\cdot)$ is the indicator function and $\tau_q$ is the confidence threshold. In practice, we usually choose a low value for $\tau_q$ so that most of the unlabeled samples can be utilized. Different from the traditional consistency regularization technique, we use $\bm{\widetilde{q}}_i$ instead of the predictions $\bm{q}^w_i$ on the weakly augmented samples as supervision. In this way, the generation and utilization of pseudo-labels can be disentangled to alleviate the accumulation of confirmation bias. 

As the open-set targets are produced by the closed-set and the multi-binary classifiers, we need to further optimize the two classifiers to obtain better open-set targets. In fact, optimizing the open-set classifier via pseudo-labeling leads to more discriminative features in the projection space and improves the performance of the multi-binary classifier at the same time. Then, for the closed-set classifier, we propose a double filtering strategy to select high-quality seen-class pseudo-labels of inliers: 
\vspace{-2mm}
\begin{equation}
\label{eq:loss-u}
    \mathcal{L}_{ui}(\mathcal{U})=\frac{1}{\mu B}\sum_{i=1}^{\mu B} \mathcal{F}(\bm{u}_i)\cdot \text{H}(\widetilde{\bm{p}}_i,\bm{p}^s_i).
\end{equation} 
$\mathcal{F}(\cdot)$ is the filtering function, which is defined as $\mathcal{F}(\bm{u}_i) = \mathbbm{1}(\max_k(\widetilde{p}_{i,k})>\tau_p)\cdot \mathbbm{1}(\mathcal{S}_i < 0.5)$,
where $\tau_p$ is another confidence threshold. We use $\mathcal{S}_i$ to exclude the likely outliers and use $\tau_p$ to ignore incorrect pseudo-labels of inliers. As these temporarily excluded samples have been utilized by the open-set classifier, the true inliers will be gradually involved in the training, which prevents IOMatch from falling into the same issue as the previous OSSL methods.

The overall optimization objective of IOMatch is consistent through the training, which is defined as 
\vspace{-1mm}
\begin{equation}
\label{eq:loss-overall}
    \mathcal{L}_{overall} = \mathcal{L}_s + \lambda_{mb}\mathcal{L}_{mb} + \lambda_{ui}\mathcal{L}_{ui}
 + \lambda_{op}\mathcal{L}_{op},
\end{equation}
where $\lambda_{mb}$, $\lambda_{ui}$, and $\lambda_{op}$ are the weights of each learning objective, respectively. As these learning objectives are all cross-entropy losses\footnote{The multi-binary loss $\mathcal{L}_{mb}$ can be viewed as the combination of two binary cross-entropy losses.} with the same order of magnitude, we can simply set $\lambda_{mb}=\lambda_{ui}=\lambda_{op}=1$.
In Algorithm \ref{alg:algorithm}, we present the detailed optimization procedure in every training iteration. Different from the existing OSSL methods based on the detect-and-filter strategy, IOMatch is an one-stage framework which omits a sensitive hyperparameter, \ie, the number of epochs for pre-training the outlier detector. Using only cross-entropy losses, IOMatch is much easier to implement than the methods equipped with complex contrastive learning objectives. From these aspects, IOMatch shows remarkable simplicity.

\subsection{Inference}
The well trained encoder, projector, and classifiers of IOMatch will be used for inference. For the closed-set classification task, the closed-set classifier is employed to obtain $\bm{p}_t=\phi(f(\bm{x}_t))$ and assign the test sample $\bm{x}_t$ to the $\hat{y}_t$-th seen class, where $\hat{y}_t=\argmax_k(p_{t,k})\in \{1,\dots,K\}$. For the open-set classification task that regards all unseen-class outliers as a single new class, we consider the open-set probability distribution produced by the open-set classifier, \ie, $\bm{q}_t=\psi(g(f(\bm{x_t})))$. The open-set prediction is given by $\hat{y}_t=\argmax_k(q_{t,k})\in \{1,\dots,K+1\}$. In fact, $\bm{q}_t$ can also be used for the closed-set task by ignoring its last item, while we still use $\bm{p}_t$ to be consistent with other methods.

\subsection{Connections to Existing Methods}
Although both employ the multi-binary classifier for unseen-class detection, IOMatch distinguishes itself from the previous OpenMatch \cite{saito2021openmatch} in the following aspects: (1) In IOMatch, we optimize the closed-set classifier and the multi-binary classifier in different feature spaces to mitigate mutual interference. (2) All the network modules in IOMatch are simultaneously optimized, without an extra pre-training stage for the multi-binary classifier. (3) We adopt a novel unified paradigm for jointly utilizing inliers and outliers, which is totally different from the conventional detect-and-exclude strategy.

Compared to the standard SSL methods, like FixMatch \cite{sohn2020fixmatch}, IOMatch can properly utilize the outliers to mitigate their negative affects on pseudo-labeling and even be able to achieve additional performance gains from them. Moreover, IOMatch is a general SSL framework that also performs well in the standard SSL setting. For standard SSL tasks, IOMatch can utilize the low-confidence inliers (as a kind of ``outliers''), which will be ignored by FixMatch. It will yield significant performance improvements, especially when labels are scarce.

\section{Experiments}
\subsection{Experimental Setup}
We construct the open-set SSL benchmarks using public datasets, CIFAR-10/100 \cite{krizhevsky2009learning} and ImageNet \cite{deng2009imagenet}. We adopt a similar manner to \cite{saito2021openmatch} for splitting seen and unseen classes. We conduct experiments with varying class splits and varying labeled set sizes in order to cover various open-set SSL settings. Both the closed-set and open-set performance of methods are evaluated.

\textbf{Baselines.} For standard SSL methods, we focus on the latest state-of-the-arts, including MixMatch\cite{berthelot2019mixmatch}, ReMixMatch \cite{berthelot2019remixmatch}, FixMatch \cite{sohn2020fixmatch}, CoMatch \cite{li2021comatch}, FlexMatch \cite{flexmatch}, SimMatch \cite{zheng2022simmatch} and FreeMatch \cite{wang2022freematch}. We exclude earlier deep SSL methods \cite{sajjadi2016regularization,laine2016temporal,tarvainen2017mean,miyato2018virtual} because  these methods perform worse than a model trained only with labeled data on OSSL tasks \cite{oliver2018realistic,guo2020safe,chen2020semi}. 
For open-set SSL methods, we consider the published works, including UASD \cite{chen2020semi}, DS$^3$L \cite{guo2020safe}, MTCF \cite{yu2020multi}, T2T \cite{huang2021trash}, OpenMatch \cite{saito2021openmatch} and SAFE-STUDENT \cite{he2022safe}.

\textbf{Closed-Set Evaluation.} In this work, we mainly consider the closed-set classification accuracy on the test data from seen classes only, which measures the ability of models to utilize open-set unlabeled data for helping seen-class classification. We follow USB \cite{usb2022} to report the best results of all epochs to avoid unfair comparisons caused by different convergence speeds. Each task is conducted with three different random seeds and the results are expressed as mean values with standard deviation. \par
\textbf{Open-Set Evaluation.} For open-set SSL methods, we additionally evaluate their classification performance on open-set test data including both seen and unseen classes. In testing, we regard all unseen classes as a single new class, \ie, the (\textit{K}+1)-th class. Considering the open-set test data can be extremely class-imbalanced, since the number of outliers is much larger than that of inliers from each seen class, we adopt Balanced Accuracy (BA) \cite{brodersen2010balanced} as the open-set classification accuracy, which is defined as
\begin{equation}
  BA = \frac{1}{K+1}\sum_{k=1}^{K+1} {Recall}_k, 
\end{equation}
where ${Recall}_k$ is the recall score of the $k$-th class. For each method, the evaluation uses its best checkpoint model in terms of the closed-set performance.

\textbf{Fairness of Comparisons.} We have taken utmost care to ensure fair comparisons in our evaluation. Firstly, we create a unified test bed using the USB codebase \cite{usb2022}. For the standard SSL methods, we follow the re-implementations provided by USB as they yield better results than the published ones under the standard SSL setting. As for the previous open-set SSL methods, we incorporate their released code into our test bed. Because our experimental setup differs from those of the previous works (as ours involves fewer labels, making it more challenging), we first evaluate these re-implemented methods in their original setups and observe the results that are close to or higher than those reported in the published papers, which verifies the correctness of our re-implementations. Moreover, for the hyperparameters that are common to different methods, we make sure that they have consistent values. As for method-specific hyperparameters, we refer to the optimal values provided in their original papers. Experiments of each setting are performed using the same backbone networks, the same data splits, and the same random seeds. 

\begin{table*}[htbp]
  \centering
  \caption{Closed-set classification accuracy (\%) on the \textit{seen-class} test data of CIFAR-10/100 with varying seen/unseen class splits and labeled set sizes. We report the mean with standard deviation over 3 runs of different random seeds.}
  \scalebox{0.82}{
  \setlength{\tabcolsep}{1.25mm}{
    \begin{tabular}{ccccccccccc}
    \toprule
    \multicolumn{3}{c}{Dataset} & \multicolumn{2}{c}{CIFAR-10} & \multicolumn{6}{c}{CIFAR-100} \\
    \cmidrule(r){1-3} \cmidrule(r){4-5} \cmidrule(r){6-11} \multicolumn{3}{c}{Class split (Seen / Unseen)} & \multicolumn{2}{c}{6 / 4} & \multicolumn{2}{c}{20 / 80} & \multicolumn{2}{c}{50 / 50} & \multicolumn{2}{c}{80 / 20} \\
    \cmidrule(r){1-3} \cmidrule(r){4-5} \cmidrule(r){6-7} \cmidrule(r){8-9} \cmidrule{10-11} \multicolumn{3}{c}{Number of labels per class}    & 4     & 25    & 4     &  25   & 4     & 25    & 4     & 25 \\
    \midrule
\multicolumn{1}{c|}{\multirow{7}{*}{\rotatebox[origin=c]{90}{\textcolor{Periwinkle}{\textbf{Standard SSL}}}}} 
& MixMatch \cite{berthelot2019mixmatch} & NeurIPS'19 & 43.08\mystd{1.79}  & 63.13\mystd{0.64} & 28.13\mystd{5.06} & 51.28\mystd{1.45} & 26.97\mystd{0.46} & 56.93\mystd{0.84} & 28.35\mystd{0.83} & 53.77\mystd{0.97} \\
\multicolumn{1}{c|}{} & ReMixMatch \cite{berthelot2019remixmatch} & ICLR'20 & 72.82\mystd{1.81}  & 87.08\mystd{1.12} & 36.02\mystd{3.56} & 61.83\mystd{0.81} & 37.57\mystd{1.54} & 65.80\mystd{1.33} & 40.64\mystd{2.97} & 62.90\mystd{1.07} \\
\multicolumn{1}{c|}{} & FixMatch \cite{sohn2020fixmatch} & NeurIPS'20 & 81.58\mystd{6.63} & \underline{92.94\mystd{0.80}} & \underline{46.27\mystd{0.64}} & 66.45\mystd{0.74} & 48.93\mystd{5.05} & 68.77\mystd{0.89} & 43.06\mystd{1.21} & 64.44\mystd{0.51} \\
\multicolumn{1}{c|}{} & CoMatch \cite{li2021comatch} & ICCV'21 & \underline{86.08\mystd{1.08}} & 92.57\mystd{0.47} & 43.53\mystd{3.01} & 66.82\mystd{1.37} & 43.17\mystd{0.55} & 67.85\mystd{1.17} & 37.89\mystd{1.22} & 62.04\mystd{0.08} \\
\multicolumn{1}{c|}{} & FlexMatch \cite{flexmatch} & NeurIPS'21 & 73.34\mystd{4.42} & 86.44\mystd{3.72} & 37.93\mystd{4.49} & 62.68\mystd{2.02} & 44.10\mystd{1.88} & 68.98\mystd{0.94} & 43.44\mystd{2.40} & 64.34\mystd{0.64} \\
\multicolumn{1}{c|}{} & SimMatch \cite{zheng2022simmatch} & CVPR'22 & 79.84\mystd{4.76} & 90.07\mystd{2.44} & 36.93\mystd{5.72} & \underline{67.23\mystd{1.13}} & \underline{51.53\mystd{2.02}} & \underline{69.71\mystd{1.44}} & \underline{50.32\mystd{2.57}} & \textbf{65.68\mystd{1.43}} \\
\multicolumn{1}{c|}{} & FreeMatch \cite{wang2022freematch} & ICLR'23 & 79.26\mystd{4.11} & 92.27\mystd{0.15} & 45.18\mystd{8.36} & 64.62\mystd{0.79} & 50.26\mystd{1.92} & 68.57\mystd{0.27} & 47.34\mystd{0.57} & 64.41\mystd{0.55} \\
\midrule
\multicolumn{1}{c|}{\multirow{6}{*}{\rotatebox[origin=c]{90}{\textcolor{Periwinkle}{\textbf{Open-Set SSL}}}}} 
& UASD \cite{chen2020semi} & AAAI'20 & 35.25\mystd{1.07} & 56.42\mystd{1.34} & 29.78\mystd{4.28} & 53.78\mystd{0.67} & 29.08\mystd{1.44} & 54.24\mystd{1.10} & 26.41\mystd{2.16} & 50.33\mystd{0.62} \\
\multicolumn{1}{c|}{} & DS$^3$L \cite{guo2020safe} & ICML'20 & 39.09\mystd{1.24} & 51.83\mystd{1.06} & 19.70\mystd{1.98} & 41.78\mystd{1.45} & 21.62\mystd{0.54} & 47.41\mystd{0.61} & 20.10\mystd{0.48} & 40.51\mystd{1.02} \\
\multicolumn{1}{c|}{} & MTCF \cite{yu2020multi} & ECCV'20 & 49.15\mystd{6.12} & 74.42\mystd{2.95} & 32.58\mystd{3.36} & 55.93\mystd{1.66} & 35.35\mystd{2.39} & 57.72\mystd{0.20} & 25.40\mystd{1.20} & 54.59\mystd{0.49} \\
\multicolumn{1}{c|}{} & T2T \cite{huang2021trash} & ICCV'21 & 73.89\mystd{1.55} & 85.69\mystd{1.90} & 44.23\mystd{2.27} & 65.60\mystd{0.71} & 39.31\mystd{1.16} & 68.59\mystd{0.92} & 38.16\mystd{0.59} & 63.86\mystd{0.32}\\
\multicolumn{1}{c|}{} & OpenMatch \cite{saito2021openmatch} & NeurIPS'21 & 43.63\mystd{3.26} & 66.27\mystd{1.86} & 37.45\mystd{2.67} & 62.70\mystd{1.76} & 33.74\mystd{0.38} & 66.53\mystd{0.54} & 28.54\mystd{1.15} & 61.23\mystd{0.81} \\
\multicolumn{1}{c|}{} & SAFE-STUDENT \cite{he2022safe} & CVPR'22 & 59.28\mystd{1.18} & 77.87\mystd{0.14} & 34.53\mystd{0.67} & 58.07\mystd{1.40} & 35.84\mystd{0.86} & 62.75\mystd{0.38} & 34.17\mystd{0.69} & 57.99\mystd{0.34} \\
\midrule
\rowcolor{blue!10} & \textbf{IOMatch} & \textbf{Ours}  & \textbf{89.68\mystd{2.04}} & \textbf{93.87\mystd{0.16}} & \textbf{53.73\mystd{2.12}} & \textbf{67.28\mystd{1.10}} & \textbf{56.31\mystd{2.29}} & \textbf{69.77\mystd{0.58}} & \textbf{50.83\mystd{0.99}} & \underline{64.75\mystd{0.52}} \\
\bottomrule
    \end{tabular}%
    }}
  \label{tab:closed-results}%
\end{table*}%

\begin{table*}[htbp]
  \centering
   \caption{Open-set classification balanced accuracy (\%) on the \textit{open-set} test data of CIFAR-10/100, which consist of samples from all the seen and unseen classes. We report the mean with standard deviation over 3 runs of different random seeds.}
  \scalebox{0.82}{
  \setlength{\tabcolsep}{1.25mm}{
    \begin{tabular}{ccccccccccc}
    \toprule
    \multicolumn{3}{c}{Dataset} & \multicolumn{2}{c}{CIFAR-10} & \multicolumn{6}{c}{CIFAR-100} \\
    \cmidrule(r){1-3} \cmidrule(r){4-5} \cmidrule(r){6-11} \multicolumn{3}{c}{Class split (Seen / Unseen)} & \multicolumn{2}{c}{6 / 4} & \multicolumn{2}{c}{20 / 80} & \multicolumn{2}{c}{50 / 50} & \multicolumn{2}{c}{80 / 20} \\
    \cmidrule(r){1-3} \cmidrule(r){4-5} \cmidrule(r){6-7} \cmidrule(r){8-9} \cmidrule{10-11} \multicolumn{3}{c}{Number of labels per class}    & 4     & 25    & 4     &  25   & 4     & 25    & 4     & 25 \\
    \midrule
\multicolumn{1}{c|}{\multirow{6}{*}{\rotatebox[origin=c]{90}{\textcolor{Periwinkle}{\textbf{Open-Set SSL}}}}} 
& UASD \cite{chen2020semi} & AAAI'20 & 17.10\mystd{0.32} & 36.01\mystd{0.22} & 10.50\mystd{0.83} & 26.96\mystd{0.53} & 6.92\mystd{0.55} & 32.23\mystd{0.54} & 5.77\mystd{0.21} & 27.61\mystd{1.15} \\
\multicolumn{1}{c|}{} & DS3L \cite{guo2020safe} & ICML'20 & 30.89\mystd{0.33} & 40.45\mystd{0.77} & 12.56\mystd{1.21} & 34.35\mystd{0.41} & 12.14\mystd{0.39} & 35.17\mystd{0.48} & 11.10\mystd{1.27} & 29.09\mystd{0.31} \\
\multicolumn{1}{c|}{} & MTCF \cite{yu2020multi} & ECCV'20 & 33.35\mystd{7.21} & 46.13\mystd{0.54} & 8.12\mystd{2.10} & 26.60\mystd{3.66} & 4.13\mystd{0.37} & 38.36\mystd{0.29} & 1.46\mystd{0.17} & 30.75\mystd{0.52} \\
\multicolumn{1}{c|}{} & T2T \cite{huang2021trash}  & ICCV'21 & \underline{50.57\mystd{0.38}} & \underline{61.10\mystd{0.39}} & \underline{17.17\mystd{1.37}} & 37.18\mystd{0.60} & 12.74\mystd{2.66} & 44.24\mystd{0.42} &  \underline{34.23\mystd{0.57}} & \underline{51.41\mystd{0.96}} \\
\multicolumn{1}{c|}{} & OpenMatch \cite{saito2021openmatch} & NeurIPS'21 & 14.37\mystd{0.05} & 20.35\mystd{3.50} & 8.77\mystd{2.84}& \underline{39.89\mystd{1.16}} & 7.00\mystd{0.02} & \underline{49.75\mystd{1.08}} & 6.30\mystd{0.87} & 44.83\mystd{0.62}\\
\multicolumn{1}{c|}{} & SAFE-STUDENT \cite{he2022safe} & CVPR'22 & 45.27\mystd{0.36} & 52.78\mystd{0.64} & 15.94\mystd{1.07} & 28.83\mystd{0.46} & \underline{23.98\mystd{0.88}} & 46.71\mystd{1.74} & 29.43\mystd{0.66} & 50.48\mystd{0.61} \\
\midrule
\rowcolor{blue!10} & \textbf{IOMatch} & \textbf{Ours}  & \textbf{75.08\mystd{1.92}} & \textbf{78.96\mystd{0.08}} & \textbf{45.94\mystd{1.70}} & \textbf{58.52\mystd{0.48}} & \textbf{46.36\mystd{1.93}} & \textbf{60.78\mystd{0.71}} & \textbf{39.96\mystd{0.95}} & \textbf{54.39\mystd{0.38}} \\
\bottomrule
    \end{tabular}%
    }}
  \label{tab:open-full-results}%
\end{table*}%

\subsection{Main Results}
\subsubsection{CIFAR-10 and CIFAR-100}
For CIFAR-10, we use the animal classes as seen classes and the others as unseen classes, resulting in a seen/unseen class split of $6/4$. CIFAR-100 consists of $100$ classes from $20$ super-classes. We split the super-classes into seen and unseen so that inliers and outliers will belong to different super-classes. We use the first $4$, $10$, or $16$ super-classes as seen classes, resulting in three splits of $20/80$, $50/50$, and $80/20$, respectively. For both CIFAR-10 and CIFAR-100, we randomly select $4$ or $25$ samples from the training set of each seen class as the labeled data and use the rest of the training set as the unlabeled data. We use WRN-28-2 \cite{zagoruyko2016wide} as the backbone encoder. We use an identical set of hyperparameters, which is \{$\lambda_{mb}=\lambda_{ui}=\lambda_{op}=1$, $\tau_p=0.95$, $\tau_q=0.5$, $\mu=7$, $B=64$, $N_e=256$, $N_i=1024$\}, across all tasks. $N_e$ indicates the total number of training epochs and $N_i$ is the number of iterations per epoch.

For the closed-set classification tasks, we compare the proposed IOMatch with thirteen latest standard and open-set SSL methods. For convenience, we denote the tasks on CIFAR-10 with 6 seen classes, 4 and 25 labeled samples per class as \texttt{CIFAR-6-24} and \texttt{CIFAR-6-150}, respectively. The denotations are similar for other tasks. We report the performance of the closed-set classifier to be consistent with other baselines. The results are presented in the Table \ref{tab:closed-results}. With respect to the closed-set classification accuracy, IOMatch achieves best performance in most tasks. When the class mismatch is severe and the labels are scarce, the improvements are quite remarkable. In particular, IOMatch outperforms the strongest rivals by $3.60\%$, $7.46\%$ and $4.78\%$ on \texttt{CIFAR-6-24}, \texttt{CIFAR-20-80}, and \texttt{CIFAR-50-200}, respectively. 

When more labeled samples are available and fewer unlabeled outliers exist, the performance gains of IOMatch would be smaller. The reason is that, in these less challenging tasks, the current state-of-the-art SSL method like SimMatch \cite{zheng2022simmatch}, can be relatively robust to the outliers with the help of its intricate contrastive learning objective. However, IOMatch can achieve better or comparable performance with less computation overhead. Furthermore, we intend to demonstrate that IOMatch is also compatible with these potent techniques. When coupled with the auxiliary self-supervised learning objectives \cite{berthelot2019remixmatch}, the performance of IOMatch can be further enhanced, surpassing the baselines entirely, as shown in Table \ref{tab:extension}.

Because the standard SSL methods do not have the capability to detect unseen-class outliers, we perform the open-set evaluation only with the open-set SSL methods. From the results presented in Table \ref{tab:open-full-results}, it is clear that IOMatch outperforms all the baselines by large margins. As we have discussed previously, the outlier detectors in these methods suffer severely from the label scarcity and tend to wrongly detect the vast majority of inliers as outliers, which results in the bad performance, especially when only 4 labels per class are available. 

In Table \ref{tab:open-full-results}, the outliers used for testing are similar to those processed during training, as we use the original test sets of CIFAR10/100, which include all the 10/100 classes. In order to evaluate the classification performance on the wild open-set test data, we also conduct the experiments with the test set containing foreign outliers from different datasets than CIFAR10/100. We observe that IOMatch can still achieve impressive open-set performance for this case. We present the detailed setting and corresponding results in the supplementary material.

\vspace{-3mm}

\subsubsection{ImageNet}
Following \cite{saito2021openmatch}, we choose ImageNet-30 \cite{laine2016temporal}, which is a subset of ImageNet \cite{deng2009imagenet} containing 30 classes. The first 20 classes are used as seen classes and the rest as unseen classes. For each seen class, we randomly select $1\%$ or $5\%$ of images with labels (13 or 65 samples per class, respectively) and the rest of images are unlabeled. Considering the high computation overhead, we adopt ResNet-18 \cite{He_2016_CVPR} as the backbone encoder and set \{$B=32$, $\mu=1$, $N_e=100$\} to finish the experiments in reasonable time. Other hyperparameters are kept consistent with the previous experiments on CIFAR10/100. Similarly, we denote the two tasks as \texttt{ImageNet-20-P1} and \texttt{ImageNet-20-P5}.

We select the methods achieving better performance for the complete evaluation with three different seeds. The results including closed-set and open-set classification accuracy on ImageNet-30 are presented in Table \ref{tab:in30}. On this more complex and more challenging benchmark dataset, IOMatch also demonstrates its superiority in both closed-set and open-set performance. The performance can be further improved, if we use deeper backbone networks, larger batch size, and more training epochs. Nevertheless, the current results have demonstrated the effectiveness of IOMatch when computational resources are relatively limited.

\begin{table}[tb]
  \centering
  \caption{Close-set and open-set accuracy ($\%$) on ImageNet-30 with the class split of $20/10$. We report the mean with standard deviation over 3 runs of different random seeds.}
  \scalebox{0.81}{
  \setlength{\tabcolsep}{0.85mm}{
    \begin{tabular}{cccccc}
    \toprule
    \multicolumn{2}{c}{Evaluation} & \multicolumn{2}{c}{Closed-Set} & \multicolumn{2}{c}{Open-Set} \\
    \cmidrule(r){1-2} \cmidrule(r){3-4} \cmidrule(r){5-6} \multicolumn{2}{c}{Labeled ratio} & 1\%   & 5\%   & 1\%   & 5\% \\
    \midrule
    \multicolumn{2}{c}{FixMatch} & 52.52\mystd{3.82} & 78.55\mystd{1.46} &  $-$    & $-$ \\
    \multicolumn{2}{c}{CoMatch} & 62.92\mystd{0.90} & 79.17\mystd{0.42} &  $-$    & $-$ \\
    \multicolumn{2}{c}{SimMatch} & \underline{64.15\mystd{0.94}} & \underline{80.23\mystd{0.53}} &  $-$    & $-$ \\
    \midrule
    \multicolumn{2}{c}{T2T} &   63.70\mystd{0.83}    &  78.87\mystd{0.49}     &   \underline{48.81\mystd{0.88}}    & \underline{58.51\mystd{0.41}} \\
    \multicolumn{2}{c}{OpenMatch} &    56.35\mystd{3.35}   &   73.90\mystd{1.05}    &  21.80\mystd{1.90}     & 57.25\mystd{0.76} \\
    \multicolumn{2}{c}{SAFE-STUDENT} &    58.38\mystd{2.34}   &   75.85\mystd{0.99}    &  44.08\mystd{2.09}     & 55.25\mystd{1.46} \\
    \midrule
    \rowcolor{blue!10} \multicolumn{2}{c}{\textbf{IOMatch}} & \textbf{69.18\mystd{1.68}} & \textbf{81.43\mystd{0.78}} &   \textbf{57.71\mystd{2.69}}    &  \textbf{73.94\mystd{0.99}} \\
    \bottomrule
    \end{tabular}}
}
  \label{tab:in30}%
  \vspace{0.2cm}
\end{table}%

\subsection{Ablation Analysis and Discussions}

To better understand why IOMatch can obtain state-of-the-art results on OSSL tasks, we perform extensive ablation studies on the learning objectives and corresponding hyperparameters. Besides, we present some important additional results and discuss the current design and further improvements of IOMatch in depth.

\begin{figure}[t]
  \centering
  \includegraphics[width=\linewidth]{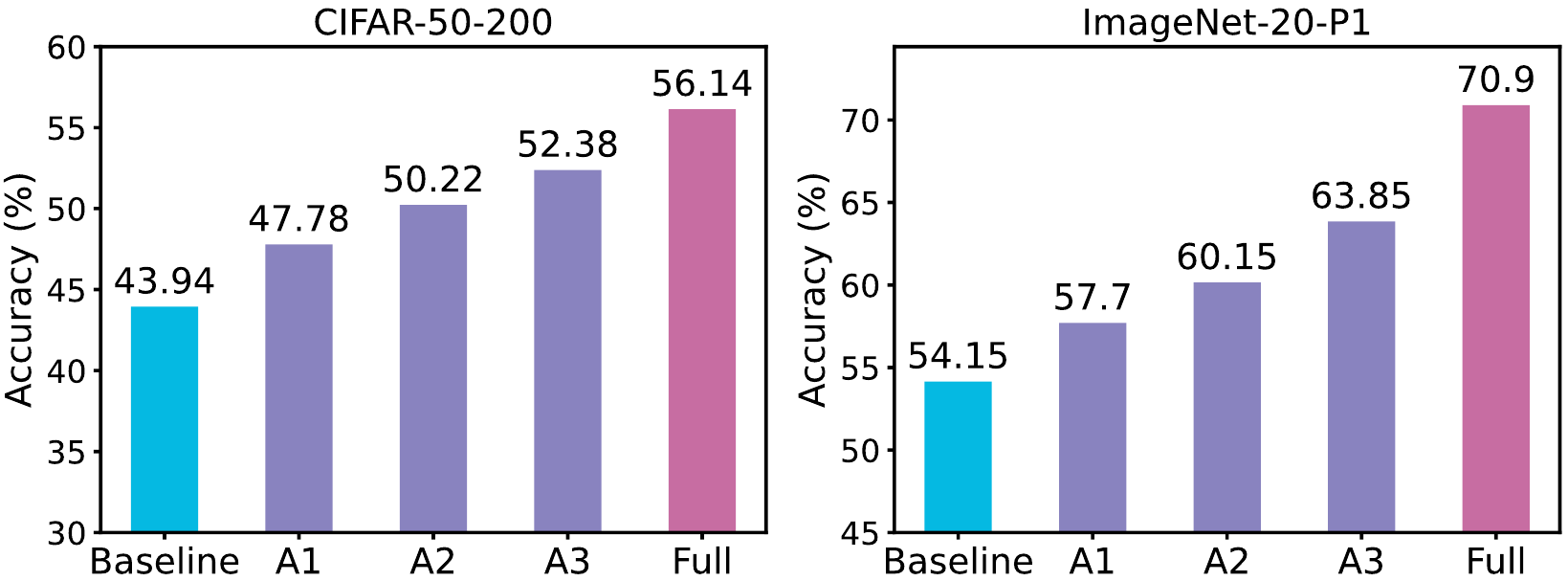}
  \caption{Ablation results on different combinations of learning objectives. "A1", "A2", and "A3" stand for the frameworks optimized with $\{\mathcal{L}_{s},\mathcal{L}_{ui}\}$, $\{\mathcal{L}_{s},\mathcal{L}_{mb},\mathcal{L}_{ui}\}$, $\{\mathcal{L}_{s},\mathcal{L}_{mb},\mathcal{L}_{op}\}$, respectively. We compare the performance with FixMatch (``Baseline'') and the full version of IOMatch (``Full'').} 
  \label{fig:ablation_objectives}
  \vspace{-2mm}
\end{figure}

\textbf{Learning Objectives.} With the standard closed-set classifier, IOMatch additionally introduces a multi-binary classifier and an open-set classifier. To examine the effects of these modules, we ablate their corresponding objectives, $\mathcal{L}_{ui}$, $\mathcal{L}_{mb}$, and $\mathcal{L}_{op}$, respectively. The results are presented in Figure \ref{fig:ablation_objectives}. Comparing ``A2'' with ``A1'', using the multi-binary classifier alone can bring some improvement as it can help to select more accurate closed-set pseudo-labels. From the results of ``A3'', we find the unsupervised training of the closed-set classifier is still important for producing better open-set targets. Most importantly, the comparisons demonstrate that the joint inliers and outliers utilization achieved by $\mathcal{L}_{op}$ is the key to the success.

\begin{figure}[t]
  \centering
  \includegraphics[width=\linewidth]{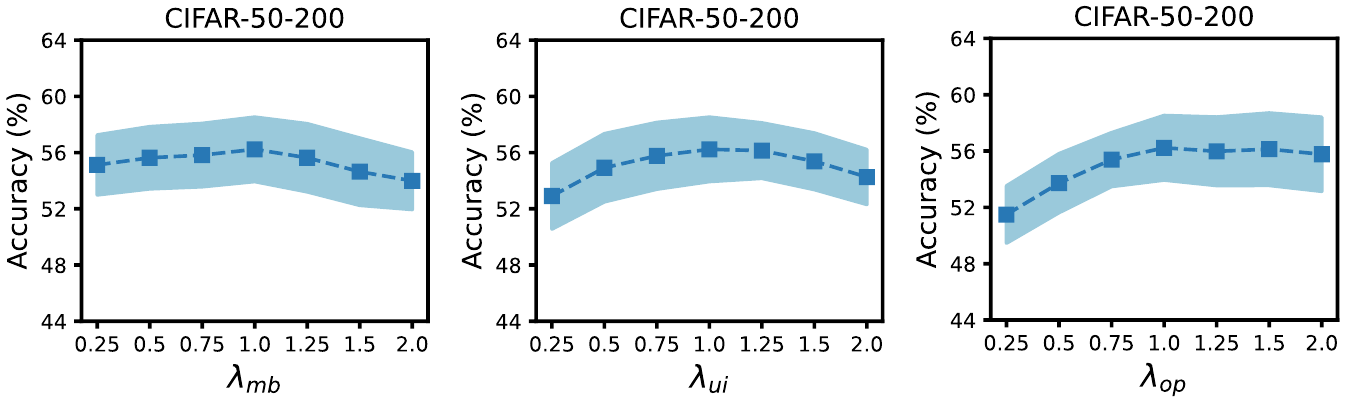}
  \vspace{-5mm}
  \caption{Performance with different values of each weight (\ie, $\lambda_{mb}$, $\lambda_{ui}$ and $\lambda_{op}$). It is shown that setting all the weights to $1$ is a simple yet appropriate choice.} 
  \label{fig:weights}
\end{figure}

\textbf{Weights of Losses.} We separately set the value of each weight (\ie, $\lambda_{mb}$, $\lambda_{ui}$ and $\lambda_{op}$) to traverse \{$0.25$, $0.5$, $0.75$, $1$, $1.25$, $1.5$, $2$\}, and control the other two weights to be $1$. And please note that we have already discussed the extreme cases where the weights are set to $0$ in the above ablation study. As shown in the Figure \ref{fig:weights}, the performance remains relatively stable when the weights are close to $1$; whereas the weights that are too small or too large may lead to performance degradation. Since the learning objectives are all cross-entropy losses with the same order of magnitude, it is reasonable to balance them with similar weights, which is well supported by the experimental observations.

\begin{figure}[t]
  \centering
  \includegraphics[width=\linewidth]{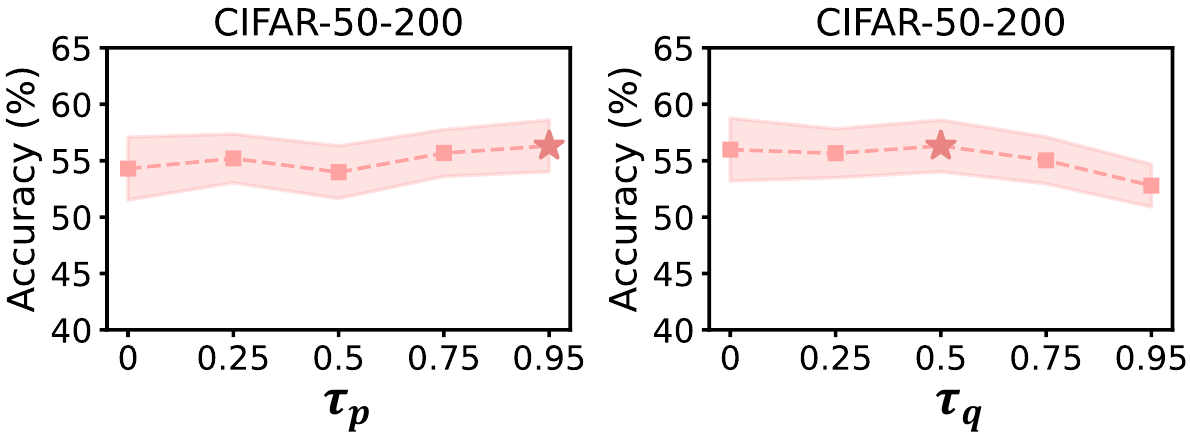}
  \vspace{-5mm}
  \caption{We vary the confidence thresholds, $\tau_p$ and $\tau_q$, respectively. The set $\{\tau_p=0.95,\tau_q=0.5\}$ gives the best performance.}
  \label{fig:thresholds}
  \vspace{-3mm}
\end{figure}

\textbf{Confidence Thresholds.} We adopt the different confidence thresholds (\ie, $\tau_p$ and $\tau_q$) for the closed-set classifier and the open-set classifier. We present the results of varying $\tau_p$ and $\tau_q$ values separately in Figure \ref{fig:thresholds}. It is shown that the performance is relatively robust to the value of $\tau_p$. Even with $\tau_p=0$, the unseen-class scores $\mathcal{S}_i$ can be used for selecting high-quality pseudo-labels alone. However, it is still helpful to choose a higher threshold (\eg, $\tau_p=0.95$ as we adopt across the tasks). As for $\tau_q$, we should choose a lower value (\eg, $\tau_q \leq 0.75$) for fully utilizing the outliers with low confidence.

\textbf{Decoupled Feature Spaces.} Different from OpenMatch \cite{saito2021openmatch}, we optimize the multi-binary classifier (and the open-set classifier) in a different feature space than the closed-set classifier, which is implemented by a projection head. We find experimentally that such design is important. For instance, if we put all the three classifiers in the same feature space (\ie, directly connected to the backbone encoder), the performance on \texttt{CIFAR-50-200} and \texttt{CIFAR-50-1250} will be reduced by about $2.2\%$ and $0.8\%$, respectively.

\begin{table}[t]
  \centering
  \caption{Closed-set classification accuracy (\%) of several methods in the standard SSL setting (presented in the column of ``SSL'') compared to the performance in the OSSL setting.}  
  \scalebox{0.85}{
  \setlength{\tabcolsep}{3.75mm}{
    \begin{tabular}{ccccc}
    \toprule
    Task  & \multicolumn{2}{c}{CIFAR-50-200} & \multicolumn{2}{c}{CIFAR-50-1250} \\
\cmidrule(r){1-1} \cmidrule(r){2-3} \cmidrule(r){4-5}    Setting & OSSL  & SSL   & OSSL  & SSL \\
    \midrule
    FixMatch & 43.94  & 45.64  & 68.92  & 72.74  \\
    SimMatch & 49.98  & 51.76  & 69.70  & \textbf{73.66} \\
    OpenMatch & 37.60  & 39.16  & 66.54  & 67.80  \\
    \midrule
    \rowcolor{purple!10} IOMatch & \textbf{56.14} & \textbf{55.94} & \textbf{69.84} & \underline{73.28}  \\
    \bottomrule
    \end{tabular}%
    }}

  \label{tab:standard}%
\end{table}%

\begin{table}[t]
  \centering
  \caption{Closed-set classification accuracy (\%) of IOMatch extended with auxiliary self-supervised learning objectives.}  
  \scalebox{0.85}{
  \setlength{\tabcolsep}{3.15mm}{
    \begin{tabular}{ccccc}
    \begin{tabular}{ccccc}
    \toprule
     Dataset & \multicolumn{4}{c}{CIFAR100} \\ 
     \cmidrule(r){1-1} \cmidrule(r){2-5}
    Class split & \multicolumn{2}{c}{50 / 50} & \multicolumn{2}{c}{80 / 20} \\ 
    \cmidrule(r){1-1} \cmidrule(r){2-3}  \cmidrule(r){4-5}
    Number of labels & 4     & 25    & 4     & 25 \\
    \midrule
    IOMatch & 56.14  & 69.84  & 49.89  & 64.28  \\
    w/ Contrastive & 57.08  & 70.80  & 50.25  & 65.92  \\
    \rowcolor{purple!10} w/ Rotation & \textbf{58.92}  & \textbf{71.54}  & \textbf{50.90}  & \textbf{66.50}  \\
    \bottomrule
    \end{tabular}%
    \end{tabular}%
    }}
  \label{tab:extension}%
  \vspace{-3mm}
\end{table}%

\textbf{Performance on Standard SSL.} We also evaluate the proposed IOMatch in the standard SSL setting where no outlier exists in unlabeled data. The results are presented in Table \ref{tab:standard}. It is shown that IOMatch is also a strong method for standard SSL, which can achieve significantly better performance when labels are scarce. When the number of labels is relatively more adequate, IOMatch can still achieve impressive performance comparable to that of advanced methods. Moreover, on the task \texttt{CIFAR-50-200}, the performance of IOMatch in the open-set setting is even better than that in the standard setting, which is made possible by the full exploitation of outliers.

\textbf{Extensions of IOMatch.} The inherent simplicity of IOMatch lends itself to the integration of other potent techniques within the framework, thereby further enhancing its performance. We explore the incorporation of self-supervised learning approaches that have exhibited remarkable effectiveness in previous methods \cite{berthelot2019remixmatch,li2021comatch,zheng2022simmatch}. Specifically, we adopt the contrastive learning objective from SimMatch \cite{zheng2022simmatch} and the rotation recognition pretext task from ReMixMatch \cite{berthelot2019remixmatch}. In the following, we introduce the implementation of the rotation recognition objective in the extended IOMatch, and the details about the contrastive learning objective can be found in the supplementary material. For each unlabeled image $\bm{u}_i$, we rotate $\bm{u}_i$ by an angle of $\angle_i$ degrees and obtain $\text{Rotate}(\bm{u}_i,\angle_i)$, where $\angle_i$ is sampled uniformly from $\angle_i \sim \{0, 90, 180, 270\}$. We add an auxiliary classifier $\theta(\cdot)$ (implemented as a fully connected layer) connected to the backbone encoder, which predicts the rotation degree among the four options, \ie, $\bm{a}=\theta(f(\text{Rotate}(\bm{u}_i,\angle_i))) \in \mathbb{R}^4$. The rotation recognition loss is defined as:
\begin{equation}
    \mathcal{L}_{rot} = \frac{1}{\mu B}\sum_{i=1}^{\mu B} \text{H}(\text{OneHot}(\angle_i), \bm{a}).
\end{equation} 
The results in Table \ref{tab:extension} demonstrate substantial performance improvements stemming from these self-supervised add-ons. It shows that our proposed IOMatch is high extensible and has great potential for enhancement.  

\textbf{Training Efficiency.} The network parameters ($15.2$M) of IOMatch are only about $3\%$ more than those ($14.7$M) of FixMatch \cite{sohn2020fixmatch}, which results in very little additional overhead. Besides, IOMatch does not require memory banks used in contrastive-based methods \cite{li2021comatch, zheng2022simmatch}, which significantly reduces the usage of GPU memory especially for large scale datasets. Therefore, IOMatch shows high training efficiency for both time and memory costs. 

\textbf{Limitations and Future Work.} Finally, we would like to discuss the limitations of the current work as well as the future directions to further improve it. In the proposed IOMatch framework, we adopt the pre-defined fixed confidence thresholds fro all classes, which could be less flexible in more complex tasks. Inspired from recent works \cite{flexmatch,wang2022freematch}, we will consider the dynamic threshold adjusting strategy for IOMatch. Besides, this work only considers the most common class space mismatch case, where the classes of labeled data form a subset of those in the unlabeled data. We will also explore other open-set scenarios, such as the intersectional mismatch, where not all labeled classes are present in the unlabeled data.

\section{Conclusion}

In this paper, we first investigate how unseen-class outliers affect the performance of the latest standard SSL methods and reveal why existing open-set SSL methods may fail when labels are extremely scarce. Inspired from the surprising fact that an unreliable outlier detector is more harmful than outliers themselves, we propose IOMatch, which adopts a novel unified paradigm for jointly utilizing open-set unlabeled data, without distinguishing exactly between inliers and outliers. Despite of its remarkable simplicity, IOMatch significantly outperforms current state-of-the-arts across various settings. We believe that the introduction of such simple but effective framework will facilitate the application of SSL methods in real-world practical scenarios.

\clearpage
{\small
\bibliographystyle{ieee_fullname}
\bibliography{egbib}

\begin{thebibliography}{10}\itemsep=-1pt

\bibitem{bachman2014learning}
Philip Bachman, Ouais Alsharif, and Doina Precup.
\newblock Learning with pseudo-ensembles.
\newblock In {\em NeurIPS}, 2014.

\bibitem{berthelot2019remixmatch}
David Berthelot, Nicholas Carlini, Ekin~D Cubuk, Alex Kurakin, Kihyuk Sohn, Han Zhang, and Colin Raffel.
\newblock Remixmatch: Semi-supervised learning with distribution alignment and augmentation anchoring.
\newblock In {\em ICLR}, 2020.

\bibitem{berthelot2019mixmatch}
David Berthelot, Nicholas Carlini, Ian Goodfellow, Nicolas Papernot, Avital Oliver, and Colin~A Raffel.
\newblock Mixmatch: A holistic approach to semi-supervised learning.
\newblock In {\em NeurIPS}, 2019.

\bibitem{brodersen2010balanced}
Kay~Henning Brodersen, Cheng~Soon Ong, Klaas~Enno Stephan, and Joachim~M Buhmann.
\newblock The balanced accuracy and its posterior distribution.
\newblock In {\em ICPR}, pages 3121--3124, 2010.

\bibitem{ChaepelleBook}
Olivier Chapelle, Bernhard Scholkopf, and Alexander Zien.
\newblock Semi-supervised learning (chapelle, o. et al., eds.; 2006) [book reviews].
\newblock {\em IEEE TNN}, 20(3):542--542, 2009.

\bibitem{chen2020simple}
Ting Chen, Simon Kornblith, Mohammad Norouzi, and Geoffrey Hinton.
\newblock A simple framework for contrastive learning of visual representations.
\newblock In {\em ICML}, pages 1597--1607, 2020.

\bibitem{chen2020semi}
Yanbei Chen, Xiatian Zhu, Wei Li, and Shaogang Gong.
\newblock Semi-supervised learning under class distribution mismatch.
\newblock In {\em AAAI}, pages 3569--3576, 2020.

\bibitem{deng2009imagenet}
Jia Deng, Wei Dong, Richard Socher, Li-Jia Li, Kai Li, and Li Fei-Fei.
\newblock Imagenet: A large-scale hierarchical image database.
\newblock In {\em ICCV}, pages 248--255, 2009.

\bibitem{duan2022rda}
Yue Duan, Lei Qi, Lei Wang, Luping Zhou, and Yinghuan Shi.
\newblock Rda: Reciprocal distribution alignment for robust semi-supervised learning.
\newblock In {\em ECCV}, pages 533--549, 2022.

\bibitem{guo2020safe}
Lan-Zhe Guo, Zhen-Yu Zhang, Yuan Jiang, Yu-Feng Li, and Zhi-Hua Zhou.
\newblock Safe deep semi-supervised learning for unseen-class unlabeled data.
\newblock In {\em ICML}, pages 3897--3906, 2020.

\bibitem{guo2022robust}
Lan-Zhe Guo, Zhi Zhou, and Yu-Feng Li.
\newblock Robust deep semi-supervised learning: A brief introduction.
\newblock {\em arXiv preprint arXiv:2202.05975}, 2022.

\bibitem{he2020momentum}
Kaiming He, Haoqi Fan, Yuxin Wu, Saining Xie, and Ross Girshick.
\newblock Momentum contrast for unsupervised visual representation learning.
\newblock In {\em CVPR}, pages 9729--9738, 2020.

\bibitem{He_2016_CVPR}
Kaiming He, Xiangyu Zhang, Shaoqing Ren, and Jian Sun.
\newblock Deep residual learning for image recognition.
\newblock In {\em CVPR}, 2016.

\bibitem{he2022safe}
Rundong He, Zhongyi Han, Xiankai Lu, and Yilong Yin.
\newblock Safe-student for safe deep semi-supervised learning with unseen-class unlabeled data.
\newblock In {\em CVPR}, pages 14585--14594, 2022.

\bibitem{hendrycks2016baseline}
Dan Hendrycks and Kevin Gimpel.
\newblock A baseline for detecting misclassified and out-of-distribution examples in neural networks.
\newblock In {\em ICLR}, 2017.

\bibitem{huang2021trash}
Junkai Huang, Chaowei Fang, Weikai Chen, Zhenhua Chai, Xiaolin Wei, Pengxu Wei, Liang Lin, and Guanbin Li.
\newblock Trash to treasure: harvesting ood data with cross-modal matching for open-set semi-supervised learning.
\newblock In {\em ICCV}, pages 8310--8319, 2021.

\bibitem{krizhevsky2009learning}
Alex Krizhevsky et~al.
\newblock Learning multiple layers of features from tiny images.
\newblock 2009.

\bibitem{laine2016temporal}
Samuli Laine and Timo Aila.
\newblock Temporal ensembling for semi-supervised learning.
\newblock In {\em ICLR}, 2017.

\bibitem{lee2018simple}
Kimin Lee, Kibok Lee, Honglak Lee, and Jinwoo Shin.
\newblock A simple unified framework for detecting out-of-distribution samples and adversarial attacks.
\newblock In {\em NeurIPS}, 2018.

\bibitem{li2021comatch}
Junnan Li, Caiming Xiong, and Steven~CH Hoi.
\newblock Comatch: Semi-supervised learning with contrastive graph regularization.
\newblock In {\em ICCV}, pages 9475--9484, 2021.

\bibitem{liu2020energy}
Weitang Liu, Xiaoyun Wang, John Owens, and Yixuan Li.
\newblock Energy-based out-of-distribution detection.
\newblock In {\em NeurIPS}, pages 21464--21475, 2020.

\bibitem{miyato2018virtual}
Takeru Miyato, Shin-ichi Maeda, Masanori Koyama, and Shin Ishii.
\newblock Virtual adversarial training: a regularization method for supervised and semi-supervised learning.
\newblock {\em IEEE TPAMI}, 41(8):1979--1993, 2018.

\bibitem{netzer2011reading}
Yuval Netzer, Tao Wang, Adam Coates, Alessandro Bissacco, Bo Wu, and Andrew~Y Ng.
\newblock Reading digits in natural images with unsupervised feature learning.
\newblock 2011.

\bibitem{oliver2018realistic}
Avital Oliver, Augustus Odena, Colin~A Raffel, Ekin~Dogus Cubuk, and Ian Goodfellow.
\newblock Realistic evaluation of deep semi-supervised learning algorithms.
\newblock In {\em NeurIPS}, 2018.

\bibitem{saito2021openmatch}
Kuniaki Saito, Donghyun Kim, and Kate Saenko.
\newblock Openmatch: Open-set consistency regularization for semi-supervised learning with outliers.
\newblock In {\em NeurIPS}, 2021.

\bibitem{saito2021ovanet}
Kuniaki Saito and Kate Saenko.
\newblock Ovanet: One-vs-all network for universal domain adaptation.
\newblock In {\em ICCV}, pages 9000--9009, 2021.

\bibitem{sajjadi2016regularization}
Mehdi Sajjadi, Mehran Javanmardi, and Tolga Tasdizen.
\newblock Regularization with stochastic transformations and perturbations for deep semi-supervised learning.
\newblock In {\em NeurIPS}, 2016.

\bibitem{sohn2020fixmatch}
Kihyuk Sohn, David Berthelot, Nicholas Carlini, Zizhao Zhang, Han Zhang, Colin~A Raffel, Ekin~Dogus Cubuk, Alexey Kurakin, and Chun-Liang Li.
\newblock Fixmatch: Simplifying semi-supervised learning with consistency and confidence.
\newblock In {\em NeurIPS}, 2020.

\bibitem{sun2021react}
Yiyou Sun, Chuan Guo, and Yixuan Li.
\newblock React: Out-of-distribution detection with rectified activations.
\newblock In {\em NeurIPS}, pages 144--157, 2021.

\bibitem{tarvainen2017mean}
Antti Tarvainen and Harri Valpola.
\newblock Mean teachers are better role models: Weight-averaged consistency targets improve semi-supervised deep learning results.
\newblock In {\em NeurIPS}, 2017.

\bibitem{van2020survey}
Jesper~E Van~Engelen and Holger~H Hoos.
\newblock A survey on semi-supervised learning.
\newblock {\em Machine Learning}, 109(2):373--440, 2020.

\bibitem{wang2022vim}
Haoqi Wang, Zhizhong Li, Litong Feng, and Wayne Zhang.
\newblock Vim: Out-of-distribution with virtual-logit matching.
\newblock In {\em CVPR}, pages 4921--4930, 2022.

\bibitem{usb2022}
Yidong Wang, Hao Chen, Yue Fan, Wang Sun, Ran Tao, Wenxin Hou, Renjie Wang, Linyi Yang, Zhi Zhou, Lan-Zhe Guo, Heli Qi, Zhen Wu, Yu-Feng Li, Satoshi Nakamura, Wei Ye, Marios Savvides, Bhiksha Raj, Takahiro Shinozaki, Bernt Schiele, Jindong Wang, Xing Xie, and Yue Zhang.
\newblock Usb: A unified semi-supervised learning benchmark for classification.
\newblock In {\em NeurIPS}, 2022.

\bibitem{wang2022freematch}
Yidong Wang, Hao Chen, Qiang Heng, Wenxin Hou, Marios Savvides, Takahiro Shinozaki, Bhiksha Raj, Zhen Wu, and Jindong Wang.
\newblock Freematch: Self-adaptive thresholding for semi-supervised learning.
\newblock In {\em ICLR}, 2023.

\bibitem{yang2021generalized}
Jingkang Yang, Kaiyang Zhou, Yixuan Li, and Ziwei Liu.
\newblock Generalized out-of-distribution detection: A survey.
\newblock {\em arXiv preprint arXiv:2110.11334}, 2021.

\bibitem{yang2021st++}
Lihe Yang, Wei Zhuo, Lei Qi, Yinghuan Shi, and Yang Gao.
\newblock St++: Make self-training work better for semi-supervised semantic segmentation.
\newblock In {\em CVPR}, 2022.

\bibitem{yang2022survey}
Xiangli Yang, Zixing Song, Irwin King, and Zenglin Xu.
\newblock A survey on deep semi-supervised learning.
\newblock {\em IEEE TKDE}, pages 1--20, 2022.

\bibitem{yu2015lsun}
Fisher Yu, Ari Seff, Yinda Zhang, Shuran Song, Thomas Funkhouser, and Jianxiong Xiao.
\newblock Lsun: Construction of a large-scale image dataset using deep learning with humans in the loop.
\newblock {\em arXiv preprint arXiv:1506.03365}, 2015.

\bibitem{yu2020multi}
Qing Yu, Daiki Ikami, Go Irie, and Kiyoharu Aizawa.
\newblock Multi-task curriculum framework for open-set semi-supervised learning.
\newblock In {\em ECCV}, pages 438--454, 2020.

\bibitem{zagoruyko2016wide}
Sergey Zagoruyko and Nikos Komodakis.
\newblock Wide residual networks.
\newblock In {\em BMVC}, 2016.

\bibitem{flexmatch}
Bowen Zhang, Yidong Wang, Wenxin Hou, Hao Wu, Jindong Wang, Manabu Okumura, and Takahiro Shinozaki.
\newblock Flexmatch: Boosting semi-supervised learning with curriculum pseudo labeling.
\newblock In {\em NeurIPS}, 2021.

\bibitem{Zhao_2022_CVPR}
Zhen Zhao, Luping Zhou, Yue Duan, Lei Wang, Lei Qi, and Yinghuan Shi.
\newblock Dc-ssl: Addressing mismatched class distribution in semi-supervised learning.
\newblock In {\em CVPR}, pages 9757--9765, 2022.

\bibitem{zheng2022simmatch}
Mingkai Zheng, Shan You, Lang Huang, Fei Wang, Chen Qian, and Chang Xu.
\newblock Simmatch: Semi-supervised learning with similarity matching.
\newblock In {\em CVPR}, pages 14471--14481, 2022.

\bibitem{zhu2022crossmatch}
Ronghang Zhu and Sheng Li.
\newblock Crossmatch: Cross-classifier consistency regularization for open-set single domain generalization.
\newblock In {\em ICLR}, 2022.

\end{thebibliography}
}

\clearpage
\onecolumn
\appendix

\newcommand{\custommaketitle}[2]{%
   \newpage
   \null
   \vskip .375in
   \begin{center}
      {\Large \bfseries #1 \par}
      \vspace*{24pt}
      {\large
      \lineskip .5em
      \begin{tabular}[t]{c}
         #2
      \end{tabular}
      \par}
      \vskip .5em
      \vspace*{12pt}
   \end{center}
}

\custommaketitle{IOMatch: Simplifying Open-Set Semi-Supervised Learning \\ with Joint Inliers and Outliers Utilization \\ \textit{Supplementary Material}}{}
\vspace{-2em}
\thispagestyle{empty}

\section{Open-Set Semi-Supervised Learning Setting}
\subsection{Class Space Mismatch} 
Open-Set Semi-Supervised Learning (OSSL) assumes that labeled and unlabeled data have different class spaces, which can be referred by the term, \textit{Class Space Mismatch}. Let $\mathcal{C}_l$ and $\mathcal{C}_u$ be the class sets of labeled and unlabeled data. Several pioneer works \cite{oliver2018realistic, chen2020semi} assume that  $\mathcal{C}_l \not\subseteq \mathcal{C}_u$ and $\mathcal{C}_u \not\subseteq \mathcal{C}_l$, while more recent OSSL works \cite{guo2020safe,yu2020multi,huang2021trash,he2022safe}  focus on the case where $\mathcal{C}_l \subset \mathcal{C}_u$. For this point, we share a similar opinion with \cite{guo2022robust}: As it is usually much easier to collect unlabeled data than labeled data, it is more likely for unlabeled data to have more categories than labeled data. Thus, we assume $\mathcal{C}_l \subset \mathcal{C}_u$ in this work.

\textbf{Remark.} A broader concept is \textit{Class Distribution Mismatch} \cite{duan2022rda,Zhao_2022_CVPR}. If we denote the marginal class distributions of labeled and unlabeled data as $\bm{p}_l(y)$ and $\bm{p}_u(y)$, then the class distribution mismatch in SSL indicates that $p_l(y)\neq p_u(y)$. The class space mismatch can be also viewed as such a case, where $p_l(y\in \mathcal{C}_u/\mathcal{C}_l)=0\neq p_u(y\in \mathcal{C}_u/\mathcal{C}_l)$. In this work, we just focus on the class space mismatch, which is the most common and problematic case of class distribution mismatch \cite{guo2022robust}.

\subsection{Connections to Out-of-Distribution Detection}
Out-of-distribution (OOD) detection \cite{hendrycks2016baseline} aims to detect OOD samples existing in test data by assigning higher OOD scores to OOD samples than ID samples. Representative works design the OOD scores using the predicted logits and probabilities \cite{liu2020energy,sun2021react}, or using the information in feature space \cite{lee2018simple}, or combining both of them \cite{wang2022vim}. More comprehensive reviews can be found in \cite{yang2021generalized}. 

Although unseen-class outliers can be also regarded as a kind of OOD samples, OOD detection is largely different from open-set SSL in the following aspects. Firstly, OOD detection tasks usually assume that sufficient labeled ID samples are provided for training (and no OOD sample exists), which cannot be satisfied in OSSL. It is a key reason why OOD detection methods cannot be directly applied in OSSL for detecting outliers. Secondly, the main objective of OOD detection is to separate OOD samples from ID samples, which can viewed as a binary classification task. However, the motivation of OSSL is to fully exploit open-set unlabeled samples for improving the model's performance on multi-class classification tasks. Therefore, a model good at OOD detection could not perform well on ID (seen-class) classification. This is the reason why we adopt Balanced Accuracy (BA) rather than AUROC, which is widely used in OOD detection, for open-set evaluation.

\section{Distribution Alignment Strategy}
For the distribution alignment (DA) strategy, we simply follow the implementation from ReMixMatch \cite{berthelot2019remixmatch}. Specifically, we maintain a running average of the model's predictions on unlabeled data, denoted by $\bm{p}_{avg}$. The marginal class distribution $\bm{p}_{mrgl}$ is estimated based on the labeled samples in training (which is the uniform distribution in our setting). Given the model's prediction $\bm{p}^w_i=\phi(f(\mathcal{T}_w(\bm{u}_i)))$ on an weakly augmented unlabeled sample $\mathcal{T}_w(\bm{u}_i)$, we scale $\bm{p}^w_i$ by the ratio $\bm{p}_{mrgl}/\bm{p}_{avg}$ and normalize the result as a valid probability distribution:
\begin{equation}
    \widetilde{\bm{p}}_i=\text{Normalize}(\bm{p}^w_i \cdot \frac{\bm{p}_{mrgl}}{\bm{p}_{avg}}),
\end{equation}
where $\text{Normalize}(\bm{p})_i=p_i/\sum_jp_j$. $\bm{p}^w_i$ is then used as the seen-class prediction for producing the unified open-set target and training the closed-set classifier via pseudo-labeling. $\bm{p}_{avg}$ is computed with the predictions over the last $128$ batches.

In practice, we find the DA strategy is effective when the number of classes is relatively large (\eg, for CIFAR-100 and ImageNet-30). However, for CIFAR-10 with fewer classes, the DA strategy may lead to performance degradation instead. The reason could be that the presence of unseen-class outliers interferes with the estimation of $\bm{p}_{avg}$. Thus, we do not apply the DA strategy in the tasks on CIFAR-10. 

\section{Extensions with Self-Supervision}
IOMatch is such a simple framework that we can easily incorporate other powerful techniques with it to further improve the performance. Recently, self-supervised learning objectives including pretext tasks \cite{gidaris2018unsupervised} and contrastive learning \cite{chen2020simple,he2020momentum} have shown strong performance in SSL \cite{berthelot2019remixmatch,li2021comatch,zheng2022simmatch}. We find experimentally that the self-supervised modules can also bring performance gains to IOMatch (see Table \ref{tab:extension} in the paper). Here we introduce the details of the extensions of IOMatch.

It is quite easy to incorporate the rotation recognition pretext task with IOMatch. For each unlabeled image $\bm{u}_i$, we rotate $\bm{u}_i$ by an angle of $\angle_i$ degrees and obtain $\text{Rotate}(\bm{u}_i,\angle_i)$, where $\angle_i$ is sampled uniformly from $\angle_i \sim \{0, 90, 180, 270\}$. We add an auxiliary classifier $\theta(\cdot)$ (implemented as a fully connected layer) connected to the backbone encoder, which predicts the rotation degree among the four options, \ie, $\bm{a}=\theta(f(\text{Rotate}(\bm{u}_i,\angle_i))) \in \mathbb{R}^4$. The rotation prediction loss is defined as:
\begin{equation}
    \mathcal{L}_{rot} = \frac{1}{\mu B}\sum_{i=1}^{\mu B} \text{H}(\text{OneHot}(\angle_i), \bm{a}).
\end{equation}

We implement the contrastive learning objective following SimMatch \cite{zheng2022simmatch}. Given the projected features of all labeled samples $\{\bm{z}_l: l \in (1,\dots, N_l)\}$ (maintained in a memory bank), the instance similarities between each unlabeled sample $\bm{u}_i$ and all labeled samples are defined as $\bm{r}_i$:
\begin{equation}
    \bm{r}_{i,l}^{w/s} = \frac{\text{exp}(\text{sim}(\bm{z}_i^{w/s},\bm{z}_l))}{\sum_{j=1}^{N_l}\text{exp}(\text{sim}(\bm{z}_i^{w/s},\bm{z}_j))},
\end{equation}
where $\text{sim}(\bm{u},\bm{v})=\bm{u}^\intercal\bm{v}/\left\|\bm{u}\right\|\left\|\bm{v}\right\|$, and $t=0.1$ is the temperature parameter. The similarity target $\widetilde{\bm{r}}$ is then generated by scaling $\bm{r}_i^w$ with $\widetilde{\bm{p}}_i$. The contrastive loss is defined as:
\begin{equation}
    \mathcal{L}_{con} = \frac{1}{\mu B}\sum_{i=1}^{\mu B} \text{H}(\widetilde{\bm{r}}_i,\bm{r}_i^s).
\end{equation}

As the above two self-supervised objectives are both standard cross-entropy losses, we can simply add them to the total loss with the weights $\mathcal{L}_{rot}=\mathcal{L}_{con}=1$. In spite of the promising results, the extensions of IOMatch introduce extra network modules (\eg, the rotation classifier and the memory bank) and thus extra training costs. It is noteworthy that, as a simple yet effective OSSL framework, IOMatch can outperform the complicated baselines on most tasks even without these extra learning objectives.

\section{Inference} 
We use the standard closed-set classifier for the inference in the closed-set classification task, in order to ensure fair comparisons with other baselines. In fact, the open-set classifier can also be used for closed-set classification by ignoring the last item of $\bm{q}_t$. We find experimentally that in this case, the predictions made by $\phi(\cdot)$ and $\psi(\cdot)$ are mostly the same.  The difference in closed-set accuracy is usually less than $0.5\%$. In the paper, we evaluate the closed-set performance using the closed-set classifier to keep consistent with other methods. However, we can just employ a single open-set classifier $\psi(\cdot)$ for both the close-set and open-set classification tasks for the sake of simplicity.

\section{Open-Set Evaluation with Foreign Outliers}
We have performed open-set evaluation with the test sets of CIFAR-10/100 (see Table \ref{tab:open-full-results} of the paper), which consist of all seen and unseen classes observed during training. In such case, unseen-class outliers in testing are similar to those in training. As the seen and unseen classes come from the same dataset, we denote them as the \textbf{intra-dataset} test data. Here we also consider the \textbf{inter-class} case where additional foreign outliers come from different datasets than CIFAR10/100. In particular, we add samples from SVHN \cite{netzer2011reading}, LSUN \cite{yu2015lsun}, and synthetic Gaussian and uniform noise images \cite{yu2020multi} as part of the testing data.

The results are shown in \ref{tab:inter-results}. Since the added foreign outliers are more dissimilar to the inliers, they are easier to identify. Therefore, the open-set accuracy on the inter-dataset test data is a little higher than that on the intra-dataset test data, while the difference is not significant.

\begin{table*}[htbp]
  \centering
    \caption{Open-set classification balanced accuracy (\%) on the \textbf{inter-dataset} open-set test data, which contain samples from different datasets than CIFAR10/100.}
  \scalebox{0.81}{
  \setlength{\tabcolsep}{1.25mm}{
    \begin{tabular}{ccccccccccc}
    \toprule
    \multicolumn{3}{c}{Dataset} & \multicolumn{2}{c}{CIFAR-10} & \multicolumn{6}{c}{CIFAR-100} \\
    \cmidrule(r){1-3} \cmidrule(r){4-5} \cmidrule(r){6-11} \multicolumn{3}{c}{Class split (Seen / Unseen)} & \multicolumn{2}{c}{6 / 4} & \multicolumn{2}{c}{20 / 80} & \multicolumn{2}{c}{50 / 50} & \multicolumn{2}{c}{80 / 20} \\
    \cmidrule(r){1-3} \cmidrule(r){4-5} \cmidrule(r){6-7} \cmidrule(r){8-9} \cmidrule{10-11} \multicolumn{3}{c}{Number of labels per class}    & 4     & 25    & 4     &  25   & 4     & 25    & 4     & 25 \\
    \midrule
\multicolumn{1}{c|}{\multirow{6}{*}{\rotatebox[origin=c]{90}{\textcolor{Periwinkle}{\textbf{Open-Set SSL}}}}} 
& UASD \cite{chen2020semi} & AAAI'20 & 18.32\mystd{0.61} & 35.78\mystd{0.22} & 11.03\mystd{0.43} & 27.35\mystd{0.33} & 7.03\mystd{0.45} & 31.94\mystd{0.74} & 5.92\mystd{0.35} & 27.83\mystd{0.85} \\
\multicolumn{1}{c|}{} & DS3L \cite{guo2020safe} & ICML'20 & 31.38\mystd{0.52} & 40.92\mystd{0.68} & 13.05\mystd{1.03} & 35.03\mystd{0.47} & 11.84\mystd{0.79} & 34.88\mystd{0.57} & 11.38\mystd{0.89} & 29.32\mystd{0.38} \\
\multicolumn{1}{c|}{} & MTCF \cite{yu2020multi} & ECCV'20 & 28.35\mystd{4.84} & 46.06\mystd{0.69} & 8.16\mystd{2.12} & 26.77\mystd{3.70} & 4.14\mystd{0.38} & 38.04\mystd{0.15} & 1.46\mystd{0.17} & 30.51\mystd{0.27} \\
\multicolumn{1}{c|}{} & T2T \cite{huang2021trash}  & ICCV'21 & \underline{51.35\mystd{1.76}} & \underline{61.78\mystd{0.89}} & \underline{17.82\mystd{1.57}} & 37.78\mystd{0.73} & 12.33\mystd{1.87} & 43.86\mystd{0.71} &  \underline{34.45\mystd{0.67}} & \underline{51.77\mystd{1.03}} \\
\multicolumn{1}{c|}{} & OpenMatch \cite{saito2021openmatch} & NeurIPS'21 & 14.37\mystd{0.05}& 20.31\mystd{3.49}& 8.77\mystd{2.83} & \underline{39.96\mystd{1.17}} & 9.97\mystd{0.37} & \underline{49.56\mystd{1.15}} & 6.31\mystd{0.88} & 44.77 \mystd{0.58} \\
\multicolumn{1}{c|}{} & SAFE-STUDENT \cite{he2022safe} & CVPR'22 & 46.37\mystd{0.61} & 54.23\mystd{0.42} & 16.31\mystd{0.88} & 29.44\mystd{0.56} & \underline{23.31\mystd{0.93}} & 46.91\mystd{1.42} & 29.52\mystd{0.55} & 50.83\mystd{0.41} \\
\midrule
\rowcolor{blue!10} & \textbf{IOMatch} & \textbf{Ours}  & \textbf{77.82\mystd{2.48}} & \textbf{82.44\mystd{0.54}} & \textbf{46.97\mystd{2.05}} & \textbf{60.30\mystd{0.99}} & \textbf{46.09\mystd{1.98}} & \textbf{60.64\mystd{0.79}} & \textbf{40.08\mystd{0.75}} & \textbf{54.57\mystd{0.30}} \\
\bottomrule
    \end{tabular}%
    }}
  \label{tab:inter-results}%
\end{table*}%

\end{document}